%% file: submission_format_fin.tex
\documentclass[a4paper,fleqn]{cas-sc}

\usepackage{times}
\usepackage{soul}
\usepackage{color}
\usepackage{url}
\usepackage[utf8]{inputenc}
\usepackage[small]{caption}
\usepackage{graphicx}
\usepackage{graphics}
\graphicspath{{figures/}{./figures/}}
\usepackage{amsmath}
\usepackage{amssymb}
\usepackage{amsthm}
\usepackage{booktabs}
\usepackage{multirow}
\usepackage{adjustbox}
\usepackage{tablefootnote}
\usepackage{longtable}
\usepackage{array}
\usepackage{calc}
\usepackage{xurl}
\usepackage{xspace}
\usepackage[authoryear,longnamesfirst]{natbib}
\usepackage{placeins}

\def\tsc#1{\csdef{#1}{\textsc{\lowercase{#1}}\xspace}}
\tsc{WGM}
\tsc{QE}
\tsc{EP}
\tsc{PMS}
\tsc{BEC}
\tsc{DE}

\makeatletter
\newenvironment{flowfigure}{\@float{figure}[!htbp]}{\end@float}
\makeatother

\colorlet{RevisionBlue}{black}
\DeclareRobustCommand{\rev}[1]{#1}
\newcommand{\RevisionInk}{}
\newcommand{\RevFigure}{}
\newcommand{\RevTable}{}
\newcommand{\RevHeading}{}
\newcommand{\RevNextEquationNumber}{}
\begin{document}
\let\WriteBookmarks\relax
\def\floatpagepagefraction{1}
\def\textpagefraction{.001}

\shorttitle{CuraView}

\shortauthors{S. Ye et~al.}

\title [mode = title]{CuraView: A Knowledge-Based Multi-Agent Framework for Patient-Grounded Medical Hallucination Detection with GraphRAG-Enhanced Evidence Verification}

\author[1]{Severin Ye}\fnmark[1]
\ead{6severin9@gmail.com}

\author[1]{Xiao Kong}\fnmark[1]
\ead{kongx7546@knu.ac.kr}

\author[2]{Xiaopeng He}\fnmark[2]
\ead{hxp19180945016@stu.wchscu.cn}

\author[3]{Guangsu Yan}\fnmark[2]
\ead{202300800569@mail.sdu.edu.cn}

\author[1,4]{Limei Peng}
\ead{auroraplm@knu.ac.kr}

\author[5]{Dongsuk Oh}\cormark[1]
\ead{inow3555@knu.ac.kr}

\address[1]{School of Computer Science and Engineering, Kyungpook National University, Daegu 41566, Republic of Korea}
\address[2]{West China Hospital, Sichuan University, Chengdu 610041, People's Republic of China}
\address[3]{School of Airspace Science and Engineering, Shandong University, Weihai 264209, People's Republic of China}
\address[4]{Shenzhen Institute for Advanced Study, University of Electronic Science and Technology of China, Guangdong, China}
\address[5]{Department of English Language and Literature, Kyungpook National University, Daegu, Republic of Korea}

\fntext[fn1]{Co-first authors with equal contribution.}
\fntext[fn2]{Co-second authors with equal contribution.}
\cortext[cor1]{Corresponding author. Tel.: +82-53-950-7155; postal address: Room 508, Graduate School Building, Kyungpook National University, 80 Daehak-ro, Buk-gu, Daegu, Republic of Korea.}

\nonumnote{E-mail addresses: 6severin9@gmail.com (S. Ye), kongx7546@knu.ac.kr (X. Kong), hxp19180945016@stu.wchscu.cn (X. He), 202300800569@mail.sdu.edu.cn (G. Yan), auroraplm@knu.ac.kr (L. Peng), inow3555@knu.ac.kr (D. Oh)}

\begin{abstract}
Medical discharge summaries require faithful synthesis of patient-specific evidence from heterogeneous electronic health records (EHRs). Although large language models (LLMs) can reduce documentation burden, they may generate clinically plausible statements that are unsupported by, or directly contradicted by, the source record. Such faithfulness hallucinations are concerning in discharge documentation because they can affect medication reconciliation, follow-up planning, and inter-provider communication. We propose \textit{CuraView}, a knowledge-based multi-agent framework for sentence-level medical hallucination detection in discharge summaries. \rev{CuraView screens sentences for insufficient support (E3) or direct contradiction (E4) within retrieved patient-specific evidence.} For each patient, it constructs a GraphRAG-enhanced knowledge base from multi-table EHR evidence and verifies generated sentences through explicit evidence retrieval, relation-aware reasoning, and structured evidence grading. The framework distinguishes four \rev{discrete} evidence levels, from support to direct contradiction, and \rev{uses a single-label taxonomy of} seven hallucination \rev{types:} diagnosis, medication, laboratory result, temporal, numerical, negation, and fabricated-fact errors. Experiments on \rev{a fixed subset of} the Discharge-Me benchmark\rev{, using generation-derived reference labels with the original discharge record available as a reference, evaluate} Qwen3-14B \rev{configurations} and \rev{two locally adapted} RAGTruth-style and QAGS-style baselines. \rev{These comparisons evaluate complete system configurations and illustrate patient-specific knowledge representation for traceable clinical-text verification.}
\end{abstract}

\begin{keywords}
medical hallucination detection \sep
GraphRAG \sep
patient-grounded verification \sep
electronic health records \sep
clinical NLP \sep
multi-agent systems \sep
interpretable evidence grading
\end{keywords}

\maketitle

\section{Introduction}
Large language models (LLMs) have shown substantial potential in clinical documentation, including the generation of discharge summaries \cite{singhal2023medpalm,singhal2023medpalm2,nori2023gpt4medical}. A discharge summary is \rev{both} a narrative record \rev{and} a clinically actionable handover document\rev{. It} communicates diagnoses, medications, laboratory findings, procedures, follow-up plans, and unresolved problems across care settings. Errors in this document may therefore affect medication reconciliation, outpatient follow-up, and subsequent clinical decision-making \cite{williams2025discharge,asgari2025clinicalsafety}. A major obstacle to safe LLM-assisted discharge documentation is faithfulness hallucination, in which a generated statement is unsupported by, or inconsistent with, the patient-specific source record \cite{ji2023hallucination}. In this setting, an incorrect medication dose, a fabricated laboratory value, or a reversed diagnostic assertion can create patient-safety risks even when the generated sentence is fluent and clinically plausible.

\begingroup\RevFigure\begin{center}
\begin{minipage}{\linewidth}
\centering
\includegraphics[width=\linewidth]{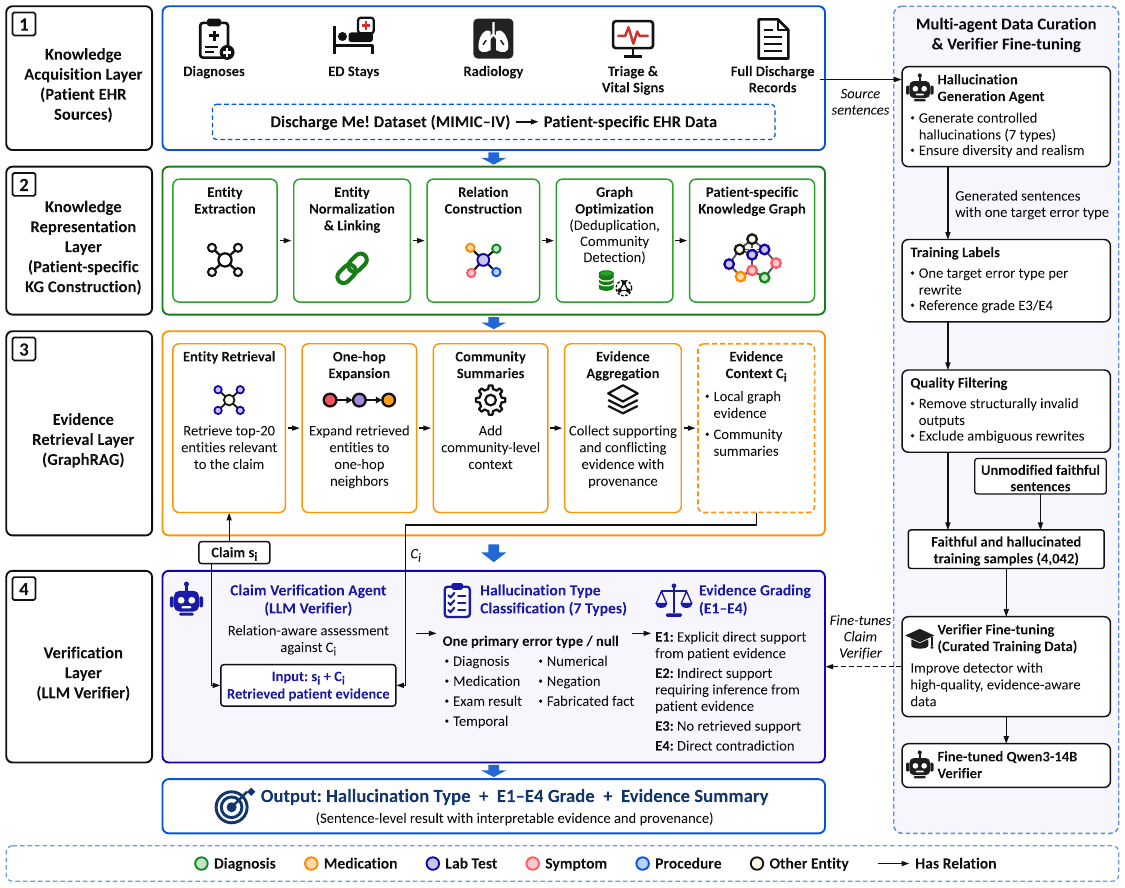}
\captionof{figure}{\rev{CuraView architecture for patient-grounded, sentence-level hallucination detection. For each candidate sentence, GraphRAG retrieves local evidence from the patient's EHR-derived knowledge graph and combines it with community summaries for relation-aware verification. The verifier assigns one primary error type or null and an E1--E4 evidence grade, with an evidence summary. A parallel data-curation branch combines faithful sentences and controlled perturbations for quality-filtered verifier fine-tuning.}}
\label{fig:system-architecture}
\end{minipage}
\end{center}\endgroup

Detecting such errors \rev{requires} patient-grounded knowledge reasoning rather than generic \rev{text classification.} The evidence required to verify a single sentence may be distributed across diagnosis codes, medication records, triage information, radiology reports, laboratory values, and free-text clinical notes. For example, verifying a statement about infection treatment may require linking a diagnosis, medication, dosage, timing, and laboratory evidence across different EHR sources. Flat retrieval methods can return relevant-looking passages, but they do not explicitly model patient-specific entities, relations, or multi-hop evidence chains. This motivates a knowledge-based architecture that organizes heterogeneous EHR evidence into structured patient-level representations and supports relation-aware verification.

Existing work has made important progress in medical hallucination evaluation, clinical text generation, and retrieval-augmented verification, but several limitations remain for discharge-summary quality control. G1: limited clinical error coverage. Many benchmarks focus on QA-style factual hallucination or document-level quality, whereas discharge summaries require sentence-level analysis of clinically meaningful error types. G2: insufficient patient-grounded evidence representation. Generic retrieval methods typically evaluate factuality against retrieved passages rather than against a coherent patient-specific knowledge base. G3: weak interpretability for clinical review. Prior systems often return binary or scalar judgments without structured evidence grades or explicit conflicting evidence, limiting their usefulness for physician-facing review. G4: limited integration of generation, detection, and data curation. Existing approaches commonly measure hallucinations but do not produce reusable, evidence-annotated training data for improving local verification models.

To address these limitations, we present CuraView, a knowledge-based multi-agent framework for medical hallucination detection with GraphRAG-enhanced evidence verification. CuraView first constructs a patient-specific knowledge graph from multi-table EHR evidence. A hallucination generation agent then creates clinically plausible sentence-level perturbations according to a seven-type clinical taxonomy. A detection agent verifies each sentence against retrieved graph evidence and assigns one of four evidence grades: E1 \rev{explicit direct} support, E2 \rev{indirect support requiring inference from patient evidence}, E3 no \rev{retrieved} support, and E4 direct contradiction. \rev{The} EHR-derived knowledge graph \rev{is} the explicit evidence layer, \rev{and} the multi-agent pipeline supports controlled data generation, evidence-grounded detection, and curated fine-tuning of a local verifier.

CuraView is positioned as a knowledge-based clinical verification system rather than a generic hallucination detector. Its core contribution \rev{is to combine} patient-specific knowledge representation, graph-structured evidence retrieval, relation-aware reasoning, and interpretable decision outputs for high-risk medical text generation. \rev{Explicit patient knowledge representation provides a shared basis for evidence retrieval and claim verification.}

The main contributions are as follows:

\begin{itemize}
\item
  Patient-grounded formulation of discharge-summary hallucination detection. We define the task as sentence-level claim verification against patient-specific EHR evidence and introduce an E1--E4 evidence grading scheme that distinguishes supported statements, unsupported claims, and direct contradictions.
\item
  Seven-type clinical hallucination taxonomy. CuraView covers diagnosis, medication, examination/laboratory result, temporal, numerical, negation, and fabricated-fact errors, enabling fine-grained analysis of clinically meaningful failure modes.
\item
  GraphRAG-based knowledge representation for EHR verification. We construct per-patient knowledge graphs from heterogeneous EHR sources and \rev{report structural changes in entity normalization and graph connectivity after domain customization. Graph-structure analysis and complete-system detection performance are reported separately.}
\item
  Multi-agent generation--detection--curation pipeline. The framework produces evidence-annotated hallucination data and uses curated fine-tuning to improve a local Qwen3-14B verifier for patient-grounded clinical hallucination detection.
\item
  Evaluation against \rev{locally adapted} reference baselines. Experiments on held-out Discharge-Me patients \rev{compare} CuraView \rev{with} RAGTruth-style and QAGS-style \rev{configurations as complete verification systems. This evaluation characterizes different complete verification systems on held-out patients.}
\end{itemize}

\section{Related Work}
\subsection{Hallucination in Medical LLMs}
Medical large language models (LLMs), including Med-PaLM \cite{singhal2023medpalm}, Med-PaLM 2 \cite{singhal2023medpalm2}, and GPT-4-based medical systems \cite{nori2023gpt4medical}, have shown strong performance on medical question answering and clinical reasoning benchmarks. However, high benchmark performance does not ensure that generated clinical documents are faithful to patient-specific source records. In clinical documentation, a generated statement must be medically plausible \rev{and} grounded in the individual patient's evidence. This distinction is especially important for discharge summaries, where unsupported or contradictory statements can affect medication reconciliation, follow-up planning, and downstream clinical decisions.

Prior work distinguishes factual hallucinations, which conflict with established world or domain knowledge, from faithfulness hallucinations, which are inconsistent with the provided source document \cite{ji2023hallucination}. CuraView focuses on faithfulness hallucinations in discharge summaries, where the verification target is not general medical knowledge but patient-specific EHR evidence. This setting requires checking whether each generated sentence is supported by, unsupported by, or contradicted by the corresponding patient's record.

Several benchmarks and studies have investigated hallucination and safety in medical text generation. Med-HALT \cite{pal2023medhalt} provides a systematic benchmark for medical hallucination evaluation, but its primary focus is medical question answering rather than patient-level document verification. Asgari et al.~\cite{asgari2025clinicalsafety} examine hallucination and omission in clinical summaries, and Williams et al.~\cite{williams2025discharge} compare physician- and LLM-generated discharge documents from a safety and quality perspective. RadFlag \cite{zhang2024radflag} addresses sentence-level hallucination detection in radiology reports, demonstrating the importance of localized error detection in clinical narratives. The Discharge-Me shared task \cite{xu2024dischargeme} and systems such as EPFL-MAKE \cite{wu2024epflmake} advance discharge-summary generation, but the task was not originally designed as a hallucination verification benchmark.

These studies establish the importance of clinical hallucination analysis, but several limitations remain for discharge-summary verification. Existing work often focuses on QA-style hallucination, document-level quality, or single-source clinical narratives. In contrast, discharge-summary verification requires sentence-level checking against heterogeneous EHR evidence, including diagnoses, medications, laboratory results, radiology reports, triage information, and free-text notes. CuraView addresses this setting by formulating hallucination detection as patient-grounded claim verification and by providing structured E1--E4 evidence grades. Table~\ref{tab:hallucination-benchmark-positioning} summarizes the positioning of CuraView relative to representative medical hallucination and clinical text generation studies.

\begin{table*}[t]
\centering
\caption{Comparison with representative medical hallucination and clinical text generation studies.}
\label{tab:hallucination-benchmark-positioning}
\scriptsize
\setlength{\tabcolsep}{3.5pt}
\renewcommand{\arraystretch}{1.12}
\begin{tabular}{p{0.20\textwidth} p{0.34\textwidth} p{0.18\textwidth} p{0.20\textwidth}}
\toprule
\textbf{Work} & \textbf{Focus} & \textbf{Granularity} & \textbf{Patient-level grounding} \\
\midrule
Med-HALT \cite{pal2023medhalt}
& Medical QA hallucination evaluation
& Question
& No \\

Asgari et al.~\cite{asgari2025clinicalsafety}
& Hallucination and omission in clinical summaries
& Sentence
& Partial \\

Williams et al.~\cite{williams2025discharge}
& Safety and quality of discharge summaries
& Document/error
& Partial \\

RadFlag \cite{zhang2024radflag}
& Radiology-specific hallucination detection
& Sentence
& Limited to reports \\

Discharge-Me \cite{xu2024dischargeme}
& Discharge-summary generation
& Section/document
& Provides EHR records \\

\textbf{CuraView}
& Patient-grounded hallucination verification
& Sentence with E1--E4 grades
& \textbf{Yes} \\
\bottomrule
\end{tabular}
\end{table*}

\subsection{Knowledge Graphs and Retrieval-Augmented Verification}

Retrieval-augmented generation (RAG) improves the ability of language models to use external information in knowledge-intensive tasks \cite{lewis2020rag}. In clinical verification, however, flat retrieval has important limitations. A discharge-summary sentence may depend on evidence distributed across multiple EHR sources, such as diagnosis codes, medication records, laboratory values, radiology reports, triage information, and free-text clinical notes. Retrieving isolated passages can provide useful context, but it does not explicitly represent how patient-specific entities, events, and relations are connected. As a result, flat retrieval may \rev{return} topically relevant passages without identifying whether they support or contradict a patient-specific claim.

Graph-based retrieval provides a more structured alternative. GraphRAG \cite{edge2024graphrag} represents document collections through entities, relations, and community-level summaries, enabling retrieval over both local and global knowledge structures. Medical Graph RAG \cite{wu2025medicalgraphrag} further demonstrates the value of graph-based retrieval in medical reasoning by connecting generated responses to trustworthy medical sources. These approaches show that graph-structured evidence can improve reasoning over complex information sources.

Nevertheless, discharge-summary verification requires a different form of grounding. Standard medical ontologies such as UMLS, SNOMED-CT, ICD, and related terminology systems provide valuable domain knowledge, but they primarily organize population-level medical concepts rather than the evidence recorded for a specific patient encounter. CuraView therefore constructs a separate knowledge graph for each patient from multi-table EHR data. This design allows the verifier to reason over patient-level entities, relations, and evidence chains. Domain-customized graph construction is \rev{designed} to reduce entity fragmentation \rev{and} improve graph connectivity. \rev{The single-case analysis below describes graph structure and source-record agreement; the output ceilings of both configurations are specified in the experimental setup.}

\subsection{Multi-Agent Systems for Clinical Verification}

Multi-agent frameworks are increasingly used in LLM-based systems because they allow complex tasks to be decomposed into specialized roles. Frameworks such as LangChain \cite{chase2023langchain} support modular orchestration of retrieval, generation, tool use, and structured output. In the medical domain, MedAgents \cite{tang2023medagents} \rev{uses role-based, multi-round collaboration and consensus for zero-shot medical question answering. AgentClinic \cite{schmidgall2026agentclinic} evaluates tool-using clinical agents in interactive multimodal diagnostic encounters that include doctor, patient, measurement, and moderator agents. These systems study collaborative reasoning or interactive clinical decision tasks; they do not evaluate sentence-level faithfulness against a patient-specific discharge-record evidence base.} In parallel, LLM-as-a-judge approaches \cite{croxford2025clinicalai,ravi2024lynx} have been used for structured evaluation of generated clinical or factual content.

CuraView differs from these systems in both task definition and evidence source. Its goal is not diagnosis support or general medical reasoning, but sentence-level quality control for discharge summaries. The framework separates three roles: hallucination generation, patient-specific evidence organization, and hallucination detection. The generation agent creates controlled sentence-level perturbations according to clinically meaningful error types. The GraphRAG module organizes each patient's EHR evidence into a structured knowledge representation. The detection agent then verifies each sentence against retrieved graph evidence and assigns an interpretable evidence grade.

This design supports both verification and data curation. The generated examples \rev{support evaluation and} provide evidence-annotated data for adapting a local verifier. CuraView \rev{thus} links generation, detection, and curated fine-tuning within a single quality-control pipeline. This integration is important for clinical settings where local deployment, interpretable output, and patient-specific evidence grounding are required.

\subsection{Positioning within Knowledge-Based Systems}

CuraView is positioned as a knowledge-based clinical verification system rather than a general medical LLM or a generic RAG pipeline. Its central contribution is the integration of explicit knowledge representation, graph-based evidence retrieval, structured evidence grading, and multi-agent data curation for clinical document verification. This design follows the broader goals of knowledge-based systems: acquiring domain-relevant knowledge, representing it in a structured form, reasoning over that representation, and producing interpretable outputs that support human decision-making.

This positioning is particularly important in \rev{healthcare, where} physician-facing quality-control tools \rev{often need more than} a binary hallucination label. Clinicians need to know whether a sentence is \rev{explicitly and directly supported by patient evidence, indirectly supported through inference from patient evidence}, unsupported \rev{in the retrieved context}, or directly contradicted by patient records. They also need access to the evidence that led to the judgment. CuraView addresses these requirements through patient-specific GraphRAG construction and E1--E4 evidence grading, providing a transparent verification layer for high-risk LLM-generated medical text.

\section{Problem Formulation}

\subsection{Definition and Scope}

This work focuses on \emph{faithfulness hallucinations} in discharge summary generation, where a generated clinical statement is unsupported by, or inconsistent with, the patient-specific source record \cite{ji2023hallucination}. Unlike general medical QA evaluation, the task is patient-specific claim verification against electronic health record (EHR) evidence rather than general medical plausibility judgment. The hallucination taxonomy and evidence grading system below define CuraView's detection scope and output structure.

For a patient \(p\), let \(E_p\) denote the available structured and unstructured EHR evidence, and let \(\{s_i\}_{i=1}^{n}\) be the sentences segmented from a generated discharge summary. For each sentence \(s_i\), GraphRAG retrieves patient-specific evidence context \(C_i \subseteq E_p\). \rev{CuraView predicts a screening label, a clinical error type, and an evidence grade from \(C_i\). E3+E4 denotes screening for unsupported or contradicted statements within this context; an E3 prediction does not establish that supporting evidence is absent from the complete patient record \(E_p\).}

\subsection{Medical Hallucination Taxonomy}

We classify medical hallucinations into seven clinically meaningful types. \rev{Three categories concern diagnosis, medication, and examination findings:} diagnosis errors such as incorrect diseases or conditions, medication errors such as incorrect drug names or dosages, and examination/laboratory result errors such as fabricated or incorrect test findings. \rev{The remaining} clinically important types \rev{are} temporal errors, where the timing or sequence of events is inconsistent; numerical errors, where vital signs or quantitative values are misstated; negation errors, where affirmative and negative clinical relations are reversed; and fabricated-fact errors, where an unsupported clinical event or fact is introduced. Representative examples for all seven types appear in Table~2.

This taxonomy differs from prior work \cite{pal2023medhalt,ji2023hallucination} in focusing on sentence-level faithfulness errors grounded in patient-specific EHR evidence rather than population-level medical knowledge.

Formally, \rev{the available hallucination labels are}
\[
\begin{aligned}
T=\{&\text{diagnosis\_error},\ \text{medication\_error},\ \text{exam\_result\_error},\\
    &\text{time\_error},\ \text{value\_error},\ \text{negation\_error},\ \text{invented\_fact}\}.
\end{aligned}
\]

\[
{\color{RevisionBlue}H(s_i,C_i)\in T\cup\{\varnothing\},}
\]
where \rev{the implemented structured output assigns at most one primary hallucination label to each sentence and uses \(\varnothing\) for a sentence judged faithful. For generated examples, the primary reference label is assigned during generation and identifies the intended error type.} \rev{This label} is paired with the evidence grading function \(g(s_i,C_i)\to\{E1,E2,E3,E4\}\) defined below.

\subsection{Evidence Grading System}
\label{sec:evidence-grading}

\rev{Both the archived and public prompts instruct the verifier to assign one of four evidence grades without a numerical support score or threshold. The archived prompts assigned E1 to support from structured records, E2 to support from narrative text, and E3 to a newly introduced detail that lacked support. The public verifier assigns E1 when a structured record or clinical note explicitly supports the claim and E2 when support requires inference from documented patient facts. It assigns E3 when the retrieved evidence neither supports nor directly contradicts the claim, and E4 when the claim directly conflicts with a documented patient fact.}

\rev{Contradictions take priority (E4), followed by direct support (E1), indirect support (E2), and no retrieved support (E3). The output specification maps E1/E2 to a negative hallucination decision with a null error type, and E3/E4 to a positive decision with one primary error type. For E4, the explanation identifies the conflicting patient fact (Figure~\ref{fig:grading-tree}).}

\rev{We report E3+E4 screening and E4 contradiction detection.} \rev{E4 captures} statements directly contradicted by \rev{explicit patient evidence}, while E3 serves as a broader screening category for \rev{statements lacking retrieved evidential support. E4 denotes direct contradiction with patient evidence. Its clinical harm severity requires separate annotation, which was not performed in this study.}

\begingroup\RevFigure\begin{flowfigure}
\begin{minipage}{\linewidth}
\centering
\includegraphics[width=150mm]{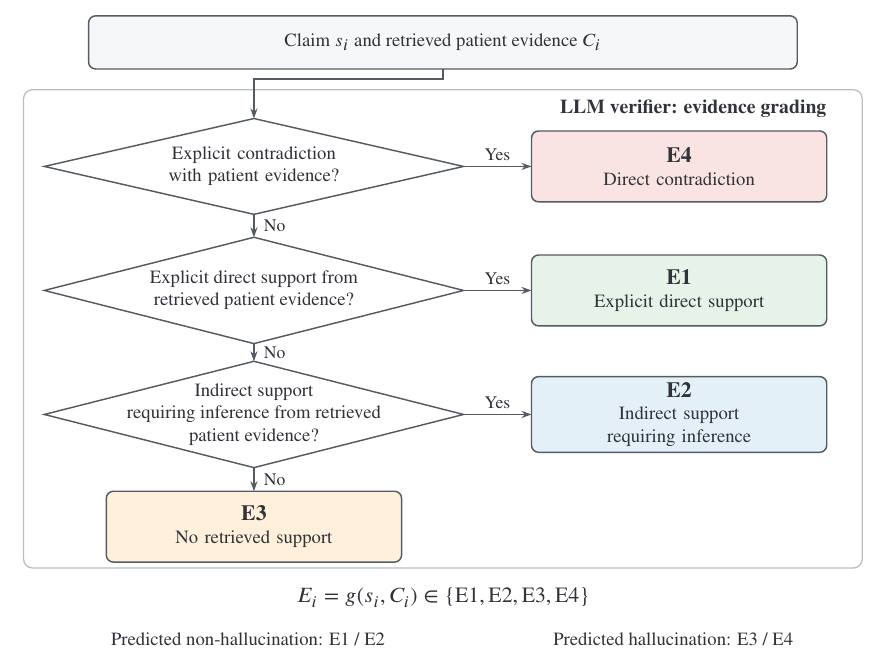}
\captionof{figure}{\rev{Discrete evidence grading using retrieved patient evidence. The verifier first checks for explicit contradiction (E4), then explicit direct support (E1), and then indirect support requiring inference (E2); otherwise, it assigns E3 for no retrieved support. E1/E2 map to predicted non-hallucination, whereas E3/E4 map to predicted hallucination.}}
\label{fig:grading-tree}
\end{minipage}
\end{flowfigure}\endgroup

\section{Methodology}

\subsection{System Architecture}

CuraView is a knowledge-based multi-agent framework for sentence-level hallucination detection in discharge summaries. The framework \rev{has} three main components. First, a hallucination generation agent creates clinically plausible sentence-level perturbations from discharge-summary sentences according to seven medical error types: diagnosis, medication, examination or laboratory result, temporal, numerical, negation, and fabricated-fact errors. Second, a GraphRAG knowledge graph module constructs a patient-specific evidence base from multi-table electronic health records (EHRs). Third, a hallucination detection agent verifies each sentence against retrieved graph evidence and produces an evidence-graded judgment.

The overall pipeline proceeds through five stages: data preparation, patient-specific graph construction, hallucination generation, hallucination detection, and evaluation. \rev{We segmented the target text into sentences before generation and detection. The verifier assessed each sentence separately.} Figure~\ref{fig:system-architecture} illustrates this workflow. \rev{The} EHR-derived graph \rev{is} the explicit knowledge layer, \rev{and} the agents perform controlled data generation and evidence-grounded verification.

\subsection{Knowledge-Based Verification Design}

CuraView treats explicit patient evidence as the primary source of truth rather than relying on the parametric memory of an LLM. \rev{We constructed a separate knowledge graph for each patient using the available EHR records listed in Table~\ref{tab:dataset-sources}.} Entities represent clinically relevant concepts such as diagnoses, medications, laboratory tests, laboratory results, vital signs, symptoms, procedures, and departments. Relations encode links among these concepts, including diagnosis assignment, medication use, test-result association, clinical indication, and department-level treatment context.

This design supports relation-aware verification. Instead of judging factuality only through surface-level similarity between a generated sentence and retrieved text, the detection agent can compare the sentence with explicit patient-level entities and evidence chains. For example, a medication statement can be checked against drug names, dosages, treatment records, and related diagnoses. This structure \rev{supports discharge-summary verification} because a single sentence may depend on multiple pieces of evidence distributed across different EHR tables.

The decision layer maps retrieved evidence into four \rev{discrete} evidence grades. \rev{E1 denotes explicit direct support from patient records, while E2 denotes indirect support requiring inference from patient evidence.} E3 indicates that a clinically relevant statement lacks \rev{support in the retrieved patient context}, while E4 indicates a direct contradiction with explicit EHR evidence. This grading scheme \rev{separates} unsupported claims from direct contradictions \rev{and distinguishes explicit direct support from indirect support.}

\subsection{Hallucination Generation Agent}

The hallucination generation agent creates controlled sentence-level perturbations for training and evaluation. The agent first screens discharge-summary sentences for clinical verifiability and rewriting suitability. Sentences that contain verifiable clinical information are then rewritten according to one of the seven hallucination types. The generation process \rev{aims to} preserve medical plausibility, maintain type diversity, and retain traceability to the source evidence.

Each generated sample contains the original sentence, the rewritten sentence, \rev{one primary} hallucination type, the evidence grade, and a short explanation of the modification. \rev{The generator is instructed to introduce one target error type per rewritten sentence.} Six error types that directly conflict with patient records are assigned E4, whereas fabricated-fact errors are assigned E3 because they introduce unsupported clinical events or facts. Table~\ref{tab:hallucination-types} \rev{illustrates the seven target error types. Each rewritten sentence is labeled with one primary error type.}

\begin{table}[t]
\centering
\caption{Representative examples of the seven hallucination types.}
\label{tab:hallucination-types}
\small
\setlength{\tabcolsep}{4pt}
\renewcommand{\arraystretch}{1.12}
\begin{tabular}{llll}
\toprule
Type & Original & Hallucination & Grade \\
\midrule
diagnosis\_error & Pneumonia & Tuberculosis & E4 \\
medication\_error & Aspirin 81 mg & Aspirin 325 mg & E4 \\
exam\_result\_error & Glucose 96 mg/dL & Glucose 196 mg/dL & E4 \\
time\_error & Admitted 1/5 & Admitted 1/15 & E4 \\
value\_error & BP 120/80 & BP 180/120 & E4 \\
negation\_error & No diabetes history & Diabetes history & E4 \\
invented\_fact & --- & Appendectomy & E3 \\
\bottomrule
\end{tabular}
\end{table}

\subsection{GraphRAG Knowledge Graph Construction}

The GraphRAG module constructs a separate knowledge graph for each patient. The graph is built from heterogeneous EHR sources and includes nine entity types: patient, diagnosis, medication, laboratory test, laboratory result, vital sign, symptom, procedure, and department. It also uses clinically meaningful relation types such as has\_diagnosis, prescribed, shows, underwent, has\_vital\_sign, tested\_by, result\_of, treated\_in, indicates, and contraindicated\_with. Table~\ref{tab:entity-types} summarizes the entity schema and representative examples.

\begin{table}[t]
\centering
\caption{Entity types used in the patient-specific EHR knowledge graph.}
\label{tab:entity-types}
\small
\setlength{\tabcolsep}{5pt}
\renewcommand{\arraystretch}{1.12}
\begin{tabular}{lll}
\toprule
Entity & \rev{Role} & Example \\
\midrule
PATIENT & \rev{Patient identity} & \texttt{Patient} \\
DIAGNOSIS & \rev{Clinical diagnosis} & \texttt{Pneumonia} \\
MEDICATION & \rev{Drug exposure} & \texttt{Aspirin} \\
LAB\_TEST & \rev{Laboratory test} & \texttt{CBC} \\
LAB\_RESULT & \rev{Test result} & \texttt{Glucose 96} \\
VITAL\_SIGN & \rev{Vital measurement} & \texttt{BP 120/80} \\
SYMPTOM & \rev{Reported symptom} & \texttt{Chest pain} \\
PROCEDURE & \rev{Procedure} & \texttt{ECG} \\
DEPARTMENT & \rev{Care setting} & \texttt{Emergency} \\
\bottomrule
\end{tabular}
\end{table}

Directly applying a general GraphRAG pipeline to EHR data can produce overly fragmented graphs. CuraView \rev{addresses this problem through} domain-customized graph construction. Laboratory measurements are normalized into clinically meaningful test panels, patient identifiers are constrained to avoid duplicate patient nodes, and synonymous clinical terms are canonicalized when possible. Table 4 \rev{reports structural results from two completed graph configurations for one longitudinal patient record; configurations and output ceilings are specified below.}

\begin{table}[t]
\centering
\caption{\rev{Observed structure of two graph configurations for one longitudinal patient record. Counts describe graph size and connectivity; the configurations use different output ceilings.}}
\label{tab:prompt-optimization}
\small
\setlength{\tabcolsep}{5pt}
\begin{tabular}{lrr}
\toprule
\rev{Measure} & \rev{Generic} & \rev{Domain} \\
\midrule
\rev{Entities} & \rev{643} & \rev{1,432} \\
\rev{Relationships} & \rev{559} & \rev{1,825} \\
\rev{Undirected connected components} & \rev{176} & \rev{7} \\
\rev{Isolated entities} & \rev{159} & \rev{6} \\
\bottomrule
\end{tabular}
\end{table}

After graph construction, communities are organized hierarchically using the Leiden algorithm \cite{edge2024graphrag}. An LLM then generates community-level summaries, which are used together with local entity evidence during sentence verification. \rev{The detection agent can thus use} fine-grained patient evidence and broader clinical context within the same patient graph.

\subsection{Hallucination Detection Agent}

The hallucination detection agent verifies each sentence \(s_i\) against the retrieved GraphRAG evidence context \(C_i\). For each sentence, the system retrieves the top entity candidates by vector similarity, expands the retrieved entities to one-hop neighbors, and integrates the resulting local evidence with community-level summaries. The final context \(C_i\) is then provided to the LLM-based verifier.

The verifier produces a structured output \(O_i = (y_i, h_i, E_i, r_i)\), where \(y_i\) is the hallucination label, \(h_i\) is the predicted hallucination type, \(E_i\) is the evidence grade, and \(r_i\) is the evidence-based reasoning. The verification process compares key clinical information in the sentence with retrieved evidence, determines whether the sentence is hallucinated, identifies the error type when applicable, and assigns one of the four evidence grades. Figure~\ref{fig:detection-pipeline} shows the detection pipeline.

\begingroup\RevFigure\begin{flowfigure}
\begin{minipage}{\linewidth}
\centering
\includegraphics[width=165mm]{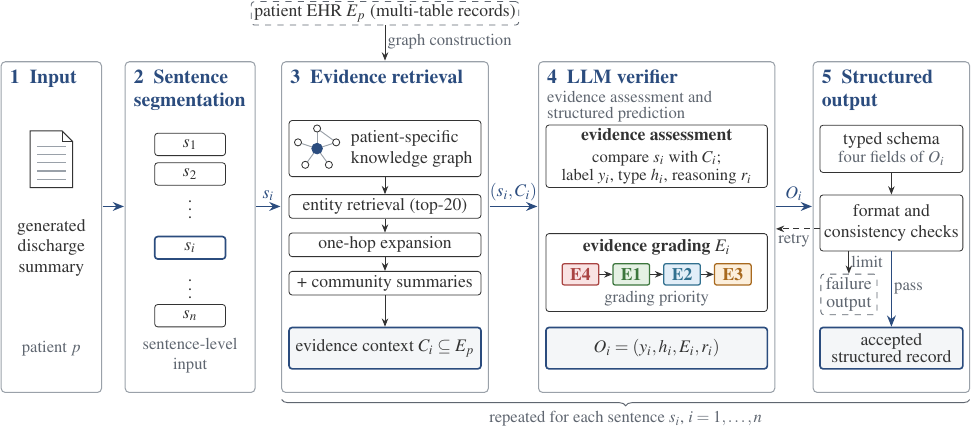}
\captionof{figure}{\rev{Sentence-level verification in CuraView. Each sentence $s_i$ of a generated discharge summary serves as both the claim and the query for retrieval over the patient-specific knowledge graph built from the EHR $E_p$; entity retrieval, one-hop expansion and community summaries yield the evidence context $C_i$. The verifier produces a structured record containing the hallucination decision, primary error type, evidence grade, and explanation. Outputs bound to a typed schema undergo format and cross-field consistency checks; failed outputs are retried, and those still failing at the retry limit are separated from accepted records (Fig.~\ref{fig:structured-output}).}}
\label{fig:detection-pipeline}
\end{minipage}
\end{flowfigure}\endgroup

To improve output reliability, CuraView uses a three-layer structured-output pipeline. First, typed schemas constrain the expected output format through LangChain's structured output interface \cite{chase2023langchain}. Second, JSON constraints and repair strategies are used to reduce format errors in local model inference. Third, parsing and validation rules check logical consistency among the hallucination label, evidence grade, and cited evidence. In particular, hallucinated sentences should be assigned E3 or E4, non-hallucinated sentences should be assigned E1 or E2, and E4 judgments should include specific conflicting evidence. This validation pipeline supports stable batch processing and improves the usability of the verifier for clinical documentation quality control \cite{liu2024structuredoutput,lu2025schemarl,geng2023grammarconstrained}, as illustrated in Figure~\ref{fig:structured-output}.

\begingroup\RevFigure\begin{flowfigure}
\begin{minipage}{\linewidth}
\centering
\includegraphics[width=150mm]{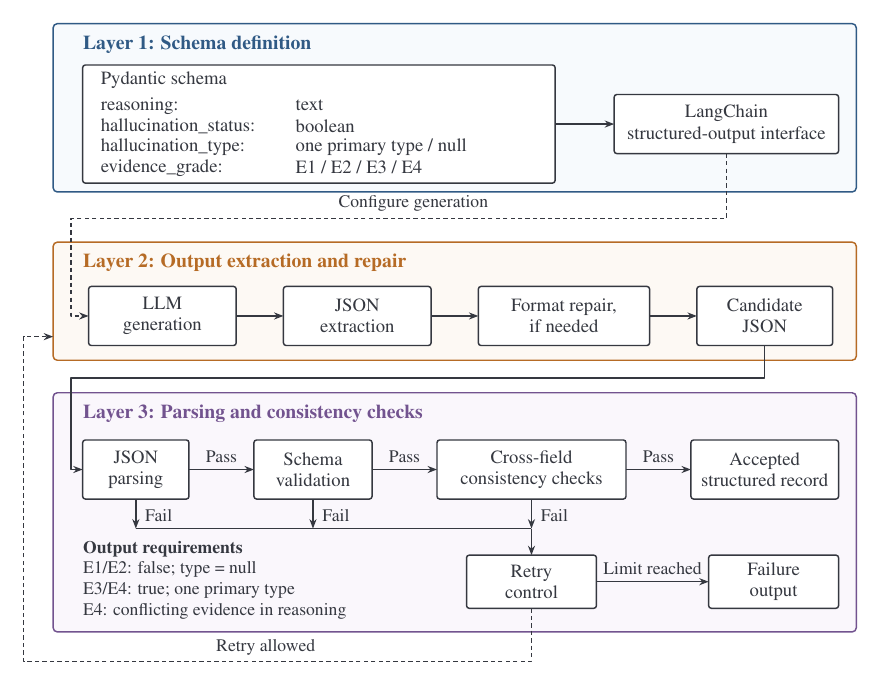}
\captionof{figure}{Three-layer structured-output \rev{processing in CuraView. Schema configuration, JSON extraction and optional format repair precede parsing, schema validation, and cross-field consistency checks. Failed checks enter retry control, with a separate failure output when the limit is reached; acceptance indicates format and field consistency, not independently verified clinical correctness.}}
\label{fig:structured-output}
\end{minipage}
\end{flowfigure}\endgroup
\FloatBarrier

\section{Experimental Setup}

\subsection{Dataset}

\subsubsection{Discharge-Me Dataset}

We evaluate CuraView on the Discharge-Me dataset \cite{xu2024dischargeme,stanfordaimi2024dischargeme}, which is derived from MIMIC-IV emergency department records collected at Beth Israel Deaconess Medical Center from 2011 to 2019. \rev{Originally designed for clinical text generation, the dataset also suits} patient-grounded hallucination verification because it provides discharge-summary targets and patient-specific source records.

\rev{The main experiment evaluates sentence-level faithfulness using unmodified sentences and controlled perturbations. Full discharge records, including the original brief hospital course where present, serve as the reference for perturbation generation and as input to graph construction. The verifier evaluates each candidate sentence against retrieved patient evidence, which can retain original target content through graph representations and community summaries. Generation records, modification explanations, and reference labels are not provided to the verifier.} Table~\ref{tab:dataset-sources} summarizes the source categories \rev{and their roles}.

\begin{center}
\begin{minipage}{\linewidth}\centering
\captionof{table}{Source categories from Discharge-Me and their role in CuraView.}
\label{tab:dataset-sources}
\small
\setlength{\tabcolsep}{4pt}
\renewcommand{\arraystretch}{1.12}
\begin{tabular}{p{0.20\linewidth}p{0.37\linewidth}p{0.28\linewidth}}
\toprule
Source & Content & Role \\
\midrule
\rev{Diagnosis records} & ICD-9/ICD-10 codes & \rev{Graph input and verifier evidence} \\
\rev{Full discharge records} & Full discharge text\rev{, which may include the original brief hospital course} & \rev{Controlled-generation reference; eligible for graph construction and verification evidence} \\
\rev{Emergency-department stays} & ED stay records & \rev{Graph input} \\
\rev{Radiology reports} & Radiology reports & \rev{Graph input and auxiliary evidence} \\
\rev{Triage records} & Triage and vital signs & \rev{Graph input and verifier evidence} \\
\bottomrule
\end{tabular}
\end{minipage}
\end{center}

\subsubsection{Data Partitioning}
\label{sec:data-partitioning}

\rev{We selected 250 patients from Discharge-Me using consecutive local identifiers 1--250. The cohort manifest contains 250 distinct subject identifiers, with the local ordering following increasing subject ID. Patients 1--200 formed the development pool for training and validation, and patients 201--250 formed the held-out test set. No subject identifier occurred in both pools.}

\begin{center}
\renewcommand{\theHtable}{main.\arabic{table}}
\captionof{table}{Experimental data partitioning and sample scale.}
\label{tab:data-partition}
\small
\setlength{\tabcolsep}{5pt}
\renewcommand{\arraystretch}{1.12}
\begin{tabular}{ll}
\toprule
Data item & Scale / notes \\
\midrule
Fixed subset & 250 patients \\
\rev{Development} pool & Patients 1--200; 4,600 samples \\
\rev{Reported training/validation ratio} & 95\% / 5\% within \rev{the development pool; see supplementary split documentation} \\
Curated training subset & Quality-filtered training samples \\
Test set & Patients 201--250; 50 patients \\
\rev{External baseline test sentences} & 1,103 sentences \\
\bottomrule
\end{tabular}
\end{center}

\rev{The study uses a subset of 250 patients to bound the cost of per-patient graph construction. The training data included faithful sentences from the original discharge records and sentences with generated hallucinations. The unfiltered development pool contained 4,600 samples, and the quality-filtered subset contained 4,042 samples. For the external baseline evaluation, we sampled 1,103 of 1,134 candidate sentences without replacement from 50 held-out patients. The sample comprised 394 generated perturbations and 709 unmodified sentences.}

\rev{Labels} are produced through the controlled \rev{hallucination-generation} process. Each rewritten sentence is paired with \rev{one primary error type} and an evidence grade. \rev{The detector receives the candidate sentence and retrieved patient context, without the generation record, modification explanation, or reference label. The benchmark labels remain generation-derived, not an independently annotated clinical gold standard.}

\subsection{Models and Baselines}

The main experiment evaluates both API-based and locally deployable model configurations. Qwen3-Plus is used as \rev{an API reference model}. We also implement two \rev{locally adapted literature-style baselines on the same held-out patient subset}. The RAGTruth-style baseline \cite{jiang2025bian} retrieves \rev{flat EHR passages and prompts} Qwen3-Plus to judge factual consistency. The QAGS-style baseline \cite{qi2025mixedcontext} generates questions from each candidate sentence, answers them \rev{from EHR source text, and compares the answers for consistency.} \rev{These workflows are local literature-inspired implementations, rather than runs of the original authors' official implementations. Differences in retrieval, verification, and output space form part of the complete-system comparison. The external comparison includes AlignScore, QAGS (no-answer variant), and FactCC (third-party checkpoint) as general-purpose factual-consistency baselines. Supplementary Table~\ref{tab:S-baselines} summarizes the baseline implementations and calibration procedures.}

\rev{For local inference, we evaluated the Qwen3-14B base model and two fine-tuned variants. Original FT used unfiltered data. Curated FT used data filtered to remove} structurally invalid outputs and ambiguous rewrites.

All fine-tuning uses LoRA under MS-SWIFT. Qwen3-14B is selected because the Qwen2.5/3 model family has shown strong performance in retrieval-oriented hallucination detection settings \cite{jiang2025bian,qi2025mixedcontext}, and task-specific aligned open-source models have been shown to be competitive in structured clinical evaluation settings \cite{croxford2025clinicalai,ravi2024lynx}. In preliminary structured-output trials, Qwen3-14B produced more reliable JSON outputs than Qwen3-8B under the same validation setting. With the full validation pipeline, the 14B model achieved stable parsing behavior for batch inference.

\subsection{Evaluation Metrics}
\label{sec:evaluation-metrics}

\rev{We evaluated predictions at the sentence level. We calculated precision and recall separately for each patient and averaged each measure across patients with equal weight. F1 was the harmonic mean of the resulting mean precision and mean recall. We matched predictions from Curated FT, Base, and AlignScore by patient identifier, sentence index, sentence text, and reference label. We estimated 95\% percentile confidence intervals for F1 differences by resampling patients with replacement 2,000 times, using the same patients for all methods in each replicate. For the external baseline evaluation and the paired comparisons, we assigned a value of zero to patient-level precision or recall when its denominator was zero. E4 F1 measures detection of statements that directly contradict explicit patient evidence; E3+E4 F1} covers both unsupported claims and direct contradictions.

\rev{We report recall on E4 contradictions to characterize missed direct conflicts with patient evidence.} Precision is also important because excessive false positives can increase alert \rev{burden.} Reporting both E4 and E3+E4 metrics \rev{characterizes direct-contradiction detection and broader screening that also includes unsupported claims.}

\subsection{Implementation Details}
\label{sec:implementation}

Local inference is performed on a single NVIDIA RTX 4090 with 24 GB VRAM. LoRA fine-tuning is performed on a single NVIDIA H200. For local inference with Qwen3-14B, we apply 8-bit quantization to fit within the 24 GB memory budget. \rev{The Qwen model configurations share the patient-level split and CuraView evaluation pipeline. The two literature-style baselines use the same held-out patient subset, each with its own retrieval, verification, and output procedures.}

For each sentence \rev{processed by CuraView}, the retrieval module selects the top 20 entity candidates by vector similarity, expands \rev{them} to one-hop neighbors, and combines local graph evidence with community-level summaries. \rev{The model-configuration comparison uses the same evaluation definitions within CuraView. The external-style baseline comparison maps each system's outputs to binary sentence decisions and compares complete systems with different evidence and decision procedures.}

\rev{The \href{https://github.com/severin-ye/CuraView/tree/d6117ca87f9418860bf4d3e4483aa73eae795bca}{CuraView repository} provides the inference code and prompt templates, with installation instructions and worked examples listed in Supplementary Tables~\ref{tab:S-materials}--\ref{tab:S-detection-materials}.}

\section{Results and Analysis}

\subsection{Main Experimental Results}

\rev{Tables~\ref{tab:q12-e4} and~\ref{tab:q12-e34} summarize sentence-level detection performance for four model configurations.} \rev{E4 contradiction detection captures} statements that directly conflict with patient records. We also report E3+E4 detection, which includes both unsupported claims and direct contradictions. This distinction allows us to evaluate the system under two clinically meaningful settings: \rev{direct-contradiction} detection and broader hallucination screening.

\begin{table}[t]
\centering
\caption{E4-level hallucination detection performance. Bold values indicate the best score in each metric column.}
\label{tab:q12-e4}
\small
\setlength{\tabcolsep}{5pt}
\renewcommand{\arraystretch}{1.12}
\begin{tabular}{lccc}
\toprule
Model & F1 (E4) & Recall (E4) & Precision (E4) \\
\midrule
Qwen3-Plus (API) & 0.791 & 0.745 & 0.842 \\
Qwen3-14B (Base) & 0.554 & 0.430 & 0.779 \\
Qwen3-14B (Original FT) & 0.803 & 0.748 & \textbf{0.867} \\
Qwen3-14B (Curated FT) & \textbf{0.831} & \textbf{0.909} & 0.765 \\
\bottomrule
\end{tabular}
\end{table}

\begin{table}[t]
\centering
\caption{Overall hallucination detection performance under the E3+E4 setting. Bold values indicate the best score in each metric column.}
\label{tab:q12-e34}
\small
\setlength{\tabcolsep}{5pt}
\renewcommand{\arraystretch}{1.12}
\begin{tabular}{lccc}
\toprule
Model & F1 (E3+E4) & Recall & Precision \\
\midrule
Qwen3-Plus (API) & 0.658 & 0.833 & 0.544 \\
Qwen3-14B (Base) & 0.777 & 0.749 & 0.808 \\
Qwen3-14B (Original FT) & 0.753 & \textbf{0.841} & 0.681 \\
Qwen3-14B (Curated FT) & \textbf{0.823} & 0.835 & \textbf{0.812} \\
\bottomrule
\end{tabular}
\par\smallskip
{\footnotesize \rev{The table reports the original summary estimates. Section~\ref{sec:q12-paired} reports the paired comparisons on sentences matched across methods.}\par}
\end{table}

\rev{The model configurations are compared using descriptive point estimates.}

\begingroup\RevFigure\begin{center}
\begin{minipage}{\linewidth}
\centering
\includegraphics[width=150mm]{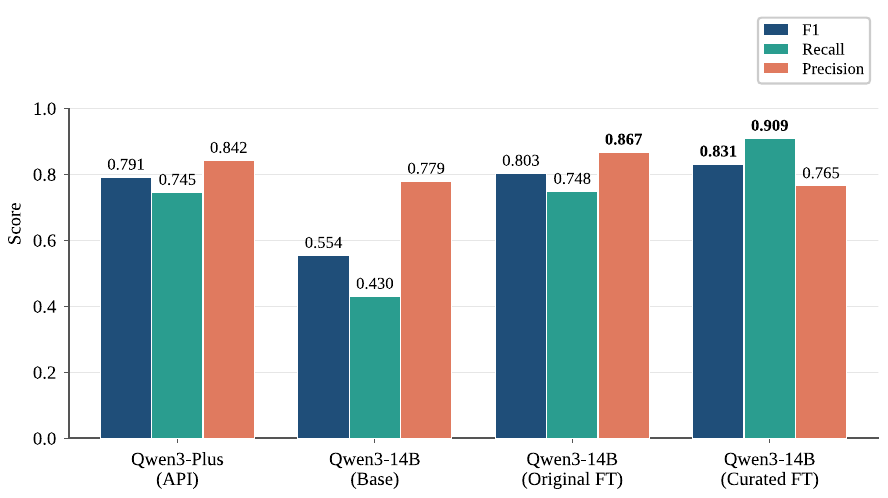}
\captionof{figure}{\rev{F1, recall and precision for E4 direct-contradiction detection across four model configurations. E4 denotes conflict with explicit patient evidence; Original FT and Curated FT denote fine-tuning on unfiltered and quality-filtered samples, respectively. Bold labels mark each metric's maximum; patient-macro averaging is defined in Section~\ref{sec:evaluation-metrics}.}}
\label{fig:result-5}
\end{minipage}
\end{center}\endgroup

\subsection{Comparison with Reference Baselines}
\label{sec:q12-paired}

\rev{We compared CuraView with two locally adapted baselines for sentence-level verification.} The RAGTruth-style \rev{method uses flat EHR retrieval and a judge model; the QAGS-style method generates questions from candidate sentences, answers them from EHRs, and compares the answers.} \rev{Because retrieval, verification and output procedures differ, the results compare complete systems.}

\begin{table*}[t]
\centering
\caption{\rev{Sentence-level performance of Curated FT and two locally adapted baselines. CuraView results are reported for E4 contradiction detection and E3+E4 hallucination detection. The baselines assess binary sentence-level consistency.}}
\small
\setlength{\tabcolsep}{5pt}
\renewcommand{\arraystretch}{1.12}
\begin{tabular}{llccc}
\toprule
Method & Evaluation setting & F1 & Recall & Precision \\
\midrule
CuraView & E4 contradiction detection & 0.831 & 0.909 & 0.765 \\
CuraView & E3+E4 hallucination detection & 0.823 & 0.835 & 0.812 \\
RAGTruth-style EHR baseline & Sentence-level consistency & 0.639 & 0.839 & 0.516 \\
QAGS-style EHR baseline & Sentence-level consistency & 0.631 & 0.850 & 0.502 \\
\bottomrule
\end{tabular}
\end{table*}

\rev{In the reported point estimates, CuraView has higher F1 than both local style baselines, while the baselines have higher recall than precision. These results characterize complete configurations that differ in retrieval, verification procedure, fine-tuning status, and output space.}

The reference baselines report \rev{binary} sentence-level consistency, whereas CuraView reports both \rev{E4 and E3+E4 settings. The comparison therefore evaluates operating configurations with different methods and label spaces.}

\begingroup\RevFigure\begin{center}
\begin{minipage}{\linewidth}
\centering
\includegraphics[width=150mm]{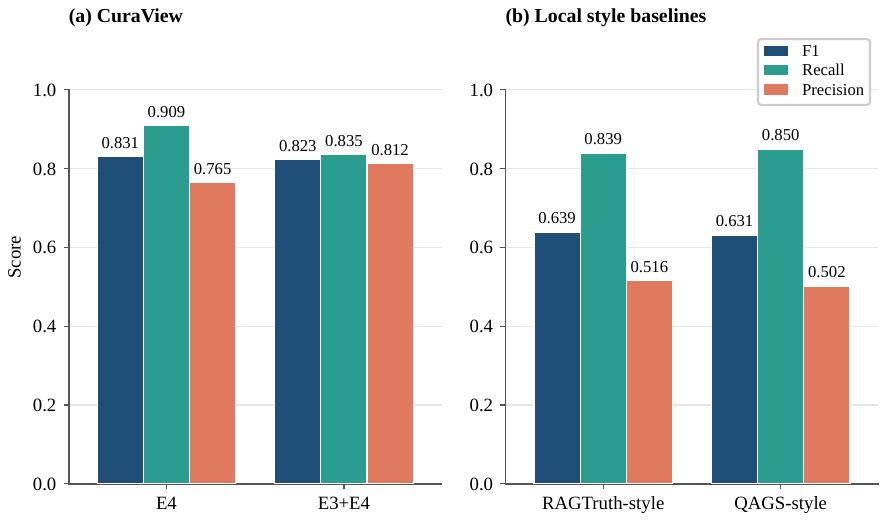}
\captionof{figure}{\rev{F1, recall and precision for (a) CuraView with curated fine-tuned Qwen3-14B under E4 and E3+E4 detection settings and (b) local RAGTruth-style and QAGS-style consistency baselines. Systems differ in retrieval, verification and label definitions; see Section~\ref{sec:evaluation-metrics} for the metric definitions.}}
\label{fig:result-6}
\end{minipage}
\end{center}\endgroup

\paragraph{\rev{External baseline evaluation.}}
\rev{We evaluated AlignScore, QAGS and FactCC on the same 1,103 test sentences. Each method used the full discharge record as its reference text. Thresholds were selected using 1,149 sentences from 50 patients in the non-test patient pool described in Section~\ref{sec:data-partitioning}. Calibration and test patient sets were disjoint. Each threshold maximized the mean per-patient hallucination F1 on the calibration set and was fixed before test evaluation. Sentences with support scores below the method-specific threshold were classified as hallucinated. The thresholds and calibration procedures are reported in Supplementary Table~\ref{tab:S-baselines}. AlignScore \cite{zha2023alignscore} uses the authors' implementation and checkpoint for CUDA inference; QAGS \cite{wang2020qags} uses the authors' model assets with the no-answer-enabled UTF-8 variant. FactCC \cite{kryscinski2020factcc} is evaluated using the third-party manueldeprada/FactCC checkpoint; the reported results apply to this implementation. Evaluation uses generation-derived labels without a separate independently annotated clinical gold standard.}

\begin{table}[t]\RevTable
\centering
\caption{\rev{External baseline performance on 1,103 test sentences from 50 patients.}}
\label{tab:new-baselines}
\small
\begin{tabular}{lrrrr}
\toprule
\rev{Method} & \rev{Precision} & \rev{Recall} & \rev{F1} & \rev{95\% CI for F1} \\
\midrule
\rev{AlignScore (CUDA)} & \rev{0.6701} & \rev{0.9746} & \rev{0.7941} & \rev{[0.7513, 0.8332]} \\
\rev{QAGS (no-answer variant)} & \rev{0.4076} & \rev{0.8974} & \rev{0.5606} & \rev{[0.5316, 0.5866]} \\
\rev{FactCC (third-party)} & \rev{0.3653} & \rev{0.9492} & \rev{0.5276} & \rev{[0.4947, 0.5546]} \\
\bottomrule
\end{tabular}
\par\smallskip
{\footnotesize \rev{We calculated 95\% percentile confidence intervals by resampling patients with replacement 2,000 times and recalculating F1 for each sample.}\par}
\end{table}

\rev{Among the three external baselines, AlignScore achieved the highest F1 (0.7941, 95\% CI 0.7513--0.8332).}

\rev{On the 616 matched sentences from 32 patients, F1 was 0.861 for Curated FT, 0.704 for Base, and 0.785 for AlignScore. The F1 difference was 0.157 (95\% CI \mbox{0.089--0.235}) for Curated FT versus Base and 0.076 (95\% CI \mbox{0.003--0.143}) for Curated FT versus AlignScore. Appendix~\ref{sec:q12-supp-eval} reports a sensitivity analysis using the binary decisions recorded by Base.}

\subsection{Ablation Studies}

\subsubsection{Fine-tuning Impact}

\rev{Table~11 compares the E4 and E3+E4 F1 scores of the Base, Original FT, and Curated FT configurations.}

\begin{table}[t]
\centering
\caption{Model evolution and fine-tuning impact. Bold values indicate the best F1 score in each metric column.}
\small
\setlength{\tabcolsep}{5pt}
\renewcommand{\arraystretch}{1.12}
\begin{tabular}{lccc}
\toprule
Configuration & F1 (E3+E4) & F1 (E4) & Change vs. base \\
\midrule
Base model (no FT) & 0.777 & 0.554 & \rev{Reference} \\
Curated fine-tuning & \textbf{0.823} & \textbf{0.831} & Improved \\
Original fine-tuning & 0.753 & 0.803 & \rev{Mixed} \\
Qwen3-Plus API & 0.658 & 0.791 & Reference \\
\bottomrule
\end{tabular}
\end{table}

\rev{Both fine-tuned configurations have higher E4 F1 point estimates than the base configuration; the curated-data configuration also has higher E3+E4 F1 than the unfiltered-data configuration. This comparison describes performance differences among the evaluated configurations and does not attribute those differences to data curation alone.}

\subsubsection{Type-wise Fine-tuning Gain Analysis}
\label{sec:curaview_typewise_gain}
\rev{We compare the base and curated fine-tuned Qwen3-14B models by hallucination type. To account for differences in baseline performance, we express the F1 change relative to the remaining headroom:}
\RevNextEquationNumber\begin{equation}
\mathrm{Gain} =
\frac{F1_{\mathrm{tuned}}-F1_{\mathrm{base}}}
{1-F1_{\mathrm{base}}}\times 100\%\mathpunct{\rev{,}}
\label{eq:curaview_typewise_gain}
\end{equation}
\rev{where $F1_{\mathrm{base}}$ and $F1_{\mathrm{tuned}}$ denote the type-wise F1 scores before and after curated fine-tuning, respectively. Table~\ref{tab:curaview_typewise_gain} reports these scores and their normalized gains, and Fig.~\ref{fig:curaview_typewise_gain} compares the gains across the seven types.}

\rev{The largest normalized gains were observed for exam-result and medication errors (43.6\% and 34.4\%, respectively). Diagnosis and numerical errors also showed positive gains. Negation errors and invented facts showed smaller positive gains, whereas temporal errors showed a slight decline ($-1.8\%$).}

\begin{table}[!htbp]
\centering
\caption{\rev{Type-wise F1 scores of the base and curated fine-tuned Qwen3-14B models and their headroom-normalized gains. Gain is defined in Section~\ref{sec:curaview_typewise_gain}. The largest gain is shown in bold.}}
\label{tab:curaview_typewise_gain}
\small
\setlength{\tabcolsep}{5pt}
\renewcommand{\arraystretch}{1.12}
\begin{tabular}{lrrr}
\toprule
Type & Base F1 & Tuned F1 & \rev{Gain (\%)} \\
\midrule
Exam result error & 0.081 & 0.482 & \textbf{+43.6} \\
Medication error & 0.381 & 0.594 & +34.4 \\
Numerical error & 0.514 & 0.600 & +17.6 \\
Diagnosis error & 0.095 & 0.278 & +20.2 \\
Negation error & 0.109 & 0.154 & +5.0 \\
Temporal error & 0.632 & 0.625 & $-1.8$ \\
Invented fact & 0.058 & 0.111 & +5.6 \\
\bottomrule
\end{tabular}
\end{table}

\begingroup\RevFigure\begin{center}
\begin{minipage}{\linewidth}
\centering
\includegraphics[width=150mm]{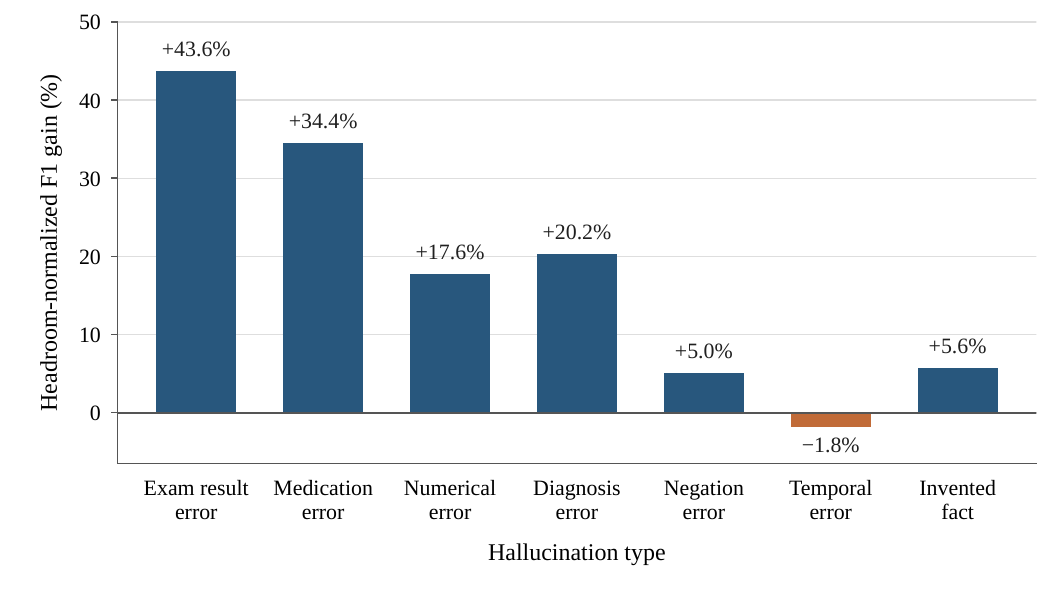}
\captionof{figure}{\rev{Headroom-normalized F1 gain across seven hallucination types after curated fine-tuning of Qwen3-14B. Each bar represents one hallucination type, with $\mathrm{Gain}=(F1_{\mathrm{tuned}}-F1_{\mathrm{base}})/(1-F1_{\mathrm{base}})\times100\%$. Positive and negative values indicate increases and decreases in type-wise F1, respectively. Corresponding F1 scores are reported in Table~\ref{tab:curaview_typewise_gain}.}}
\label{fig:curaview_typewise_gain}
\end{minipage}
\end{center}\endgroup

\subsubsection{Domain Customization Impact}

\rev{Domain customization aims to improve the organization of patient evidence. We examine graph structure for one longitudinal record (Table 4), agreement of extracted assertions with source records, coverage of fixed reference facts, and graph-retrieval-assisted detection. Connectivity describes structure, whereas the latter measures concern evidence quality and verification utility; these quantities are not interchangeable. Differences between the two construction configurations are specified under Source and configuration below.}

\begingroup\RevHeading\subsection{\rev{Exploratory Longitudinal-Record Comparison}}
\label{sec:longcase}\endgroup

\paragraph{\rev{Source and configuration.}}
\rev{The separate API exploration uses three real admissions from one patient without combining different patients. Source selection preceded outcome observation and excluded training patients and 467 previously used patients. After removal of the recognized brief-hospital-course and discharge-instruction target sections, the input contained 62,930 Unicode characters, approximately 18,001 reference-tokenizer tokens, in 18 native chunks; semantically related target content may still remain. Generic and Domain graphs were constructed from the same input using identical chunking and the same 15 entity types, distinct from the nine types listed in Table~\ref{tab:entity-types}. The Domain prompt includes medications under TREATMENT. Supplementary Table~\ref{tab:S-entity-types} compares the two entity-type inventories. Generic used a generation ceiling of 8,192 tokens. After truncation prompted an increase, the Domain configuration used 32,768 tokens and reused one normally completed extraction from the earlier 8,192-ceiling configuration. This exploration compares graph-construction configurations that differ in prompt, output-token limit, and execution history, and is separate from the Qwen3-14B fine-tuning experiment.}

\paragraph{\rev{Reference preparation and graph measures.}}
\rev{Before graph scoring, source-only preparation yielded 90 candidate reference facts. Excluding three non-atomic items and one frequency-conflict item left 86 reference facts. Each graph contributed 60 sampled records: 20 entity descriptions and 40 relationship records. Atomization was blinded to graph identity and completed before source-support assessment, yielding 137 unique assertions for Generic and 63 for Domain (Supplementary Table~\ref{tab:S-graph-sampling}). Source support is defined as \(C/T\), where \(C\) is the number of supported assertions and \(T\) is the total number of assertions; contradicted, unsupported and unresolved assertions all remain in the denominator. Strict coverage divides fully matched reference facts by the same 86 references. Partial matches and unresolved items do not count as complete, and a fact represented by several edges counts once. The references were prepared with model assistance and are provisional labels rather than clinician-adjudicated gold labels. Source-support judgments were produced by GPT-6-Astra and adjudicated by GPT-6.1-Sol. The adjudicator reviewed the initial judgments and supporting evidence without access to the graph identity key. Review covered 12 prespecified records per graph (20\%), together with unresolved or disputed items. The assertion inventories were retained, and omitted content was not added to the denominators. Further details are provided in Appendix~\ref{sec:S-graph-sampling}.}

\begingroup\RevTable\begin{table}[t]
\centering
\caption{\rev{Single-case fixed-inventory source agreement and strict reference coverage. Unresolved items remain in the denominators.}}
\label{tab:longcase-quality}
\small
\begin{tabular}{lrr}
\toprule
\rev{Measure} & \rev{Generic} & \rev{Domain} \\
\midrule
\rev{Supported assertions / all} & \rev{70/137 (51.1\%)} & \rev{57/63 (90.5\%)} \\
\rev{Contradicted assertions} & \rev{0} & \rev{1} \\
\rev{Unsupported assertions} & \rev{60} & \rev{5} \\
\rev{Unresolved assertions} & \rev{7} & \rev{0} \\
\rev{Complete references / 86} & \rev{32/86 (37.2\%)} & \rev{52/86 (60.5\%)} \\
\rev{Partially covered references} & \rev{46} & \rev{24} \\
\rev{Absent references} & \rev{7} & \rev{3} \\
\rev{Unresolved references} & \rev{1} & \rev{7} \\
\bottomrule
\end{tabular}
\end{table}\endgroup

\rev{Domain completely covered 20 more fixed reference facts than Generic, a difference of 23.3 percentage points; source support for extracted assertions differed by 39.4 percentage points. Within the prespecified review subset, assertion-boundary or omission issues were identified in 2/12 Generic records and 3/12 Domain records. Assertion support rates describe source agreement within the sampled inventories and do not estimate whole-graph clinical precision. Community reports did not undergo separate clinical review, and the 86 reference facts do not exhaust the patient record. Treating unresolved items as potentially supported gives source-support ranges of 51.1--56.2\% for Generic and 90.5--90.5\% for Domain, with possible complete-coverage ranges of 37.2--38.4\% and 60.5--68.6\%, respectively. These ranges reflect alternative judgments of unresolved items rather than sampling confidence intervals. Connectivity in Table 4 describes graph structure and measures a different aspect from source support in this table.}

\paragraph{\rev{Three retrieval conditions.}}
\rev{The retrieval comparison uses 12 constructed probes from the same patient: six source-supported statements and six single-error direct contradictions, including three initial probes and nine later additions. Flat Dense retrieves source chunks by vector similarity, whereas Generic and Domain use native graph retrieval to select and order the same source chunks. All three conditions provide source text rather than graph descriptions to the same qwen-plus detector, using the same prompt and an evidence limit of 6,000 characters. Detector inputs contain only the candidate sentence and corresponding evidence, without reference labels or generation explanations. Performance is measured as agreement between binary detector decisions and provisional source labels.}

\begingroup\RevTable\begin{table}[t]
\centering
\caption{\rev{Single-case binary detector agreement with provisional source labels; each of the three conditions produced 12 outputs, all 36 of which were format-valid. FP and FN refer to six supported and six contradictory probes per condition.}}
\label{tab:longcase-detection}
\small
\begin{tabular}{lrrrrr}
\toprule
\rev{Condition} & \rev{Correct/12} & \rev{FP} & \rev{FN} & \rev{Original 3} & \rev{Added 9} \\
\midrule
\rev{Flat Dense} & \rev{7/12} & \rev{5} & \rev{0} & \rev{1/3} & \rev{6/9} \\
\rev{Generic} & \rev{9/12} & \rev{3} & \rev{0} & \rev{1/3} & \rev{8/9} \\
\rev{Domain} & \rev{11/12} & \rev{1} & \rev{0} & \rev{2/3} & \rev{9/9} \\
\bottomrule
\end{tabular}
\end{table}\endgroup

\rev{Domain's higher binary agreement reflected fewer false positives among the six source-supported probes: 1/6 for Domain, 3/6 for Generic, and 5/6 for Flat Dense. All three configurations classified the six contradictory probes as positive. Results from one patient and 12 constructed probes do not support conclusions about population performance, clinical error prevalence, record-length effects or whole-document false-positive rates. With unequal generation ceilings and model-assisted reference labels, these observations support an exploratory comparison of the evaluated configurations rather than a causal effect of the prompt alone or clinical validity.}

\subsection{\rev{Exploratory} Meditron-7B Case Study}

\rev{To examine detector outputs} beyond the controlled perturbation setting, we apply the curated fine-tuned model to Meditron-7B-generated discharge summaries \cite{chen2023meditron,wu2024epflmake}. This \rev{exploratory} case study uses 25 MIMIC-IV patients and 154 generated sentences. \rev{CuraView flags 45 sentences, corresponding to a CuraView-predicted positive rate of 29.22\%. Here, positive means a sentence flagged as suspicious by CuraView; 45/154 describes detector outputs.}

\rev{Among sentences predicted positive by CuraView, the invented-fact label is most frequent. This distribution describes detector behavior on Meditron outputs. Without independent annotation, it is not interpreted as the true error distribution.}

\begingroup\RevFigure\begin{center}
\begin{minipage}{\linewidth}
\centering
\includegraphics[width=150mm]{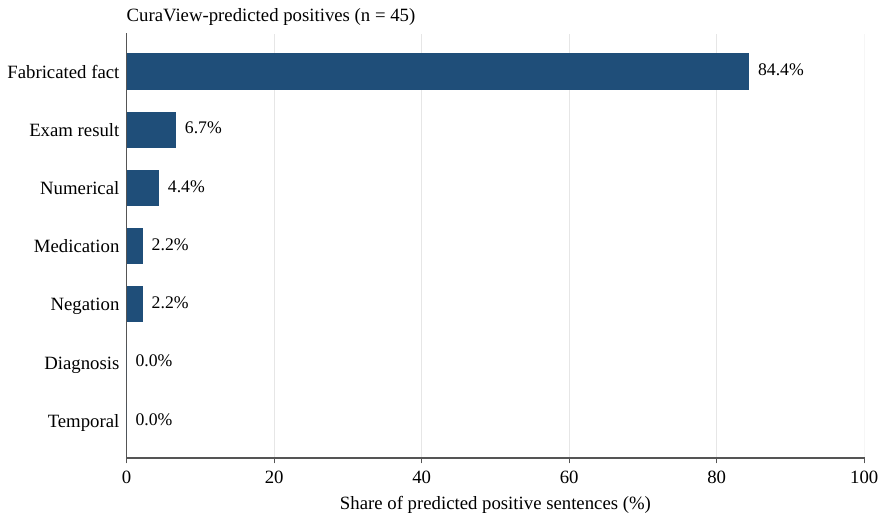}
\captionof{figure}{\rev{CuraView-predicted type distribution for Meditron-7B outputs from 25 patients and 154 generated sentences. CuraView flagged 45 sentences ($45/154=29.22\%$ predicted-positive rate), which form the denominator for all type percentages. Without independent human annotation, this distribution does not establish true hallucination prevalence, detector accuracy or generalization.}}
\label{fig:result-8}
\end{minipage}
\end{center}\endgroup

\subsection{Qualitative Analysis}

\rev{The following two schematic examples explain the labeling rules and a possible evidence-coverage failure.}

\rev{Case 1 illustrates an E4 diagnosis contradiction under the single-label task definition.} The original \rev{statement} ``Patient diagnosed with pneumonia'' was rewritten as ``Patient diagnosed with tuberculosis.'' \rev{The appropriate primary label under this schematic example is \texttt{diagnosis\_error} when the reference evidence supports pneumonia rather than tuberculosis.}

\rev{Case 2 illustrates a possible evidence-coverage failure. Consider a faithful statement about intravenous fluid replacement for dehydration. If the retrieved context omits its supporting documentation, a verifier may assign E3 despite support elsewhere in the record. This schematic scenario shows why a lack of retrieved support does not mean that the complete record establishes the statement as false.}

\rev{Further error analysis can use matched predictions and source evidence to test hypotheses of synonym mismatch and incomplete retrieved support.}

\begingroup\RevHeading\subsection{\rev{Supplementary Evaluation on Unmodified Target Documents}}
\label{sec:unmodified-docs}

\rev{A separate LoRA-tuned Qwen3-14B-Base model with Domain retrieval was evaluated on 50 complete, unmodified brief-hospital-course target documents. Sentence segmentation and the filtering rules retained 1,134 sentences, a mean of 22.68 sentences per document. The unmodified text served as the reference negative: a sentence predicted as E3 or E4 was counted as a false positive, and failed or missing outputs were reported separately rather than counted as true negatives. Of the 1,134 sentences, 986 received valid decisions, comprising 361 false positives and 625 true negatives; the false-positive rate was 36.61\% and the specificity was 63.39\%. The remaining 148 sentences were undetermined. At the document level, 80\% of the documents contained at least one false-positive sentence.}\endgroup

\section{Discussion}

\rev{Curated FT had higher E3+E4 F1 than both Base and AlignScore on the 616 matched sentences. E4 denotes direct contradiction with explicit patient evidence; it does not determine the severity of clinical harm, which was not assessed in this study.}

\rev{Patient-specific graph construction organizes information around patient entities, relations and evidence chains before verification and forms part of CuraView. Because the external-style baselines use different retrieval and verification procedures, differences in system-level point estimates cannot be attributed to graph retrieval alone. The single-case analysis further characterizes evidence organization through connectivity, source support for extracted assertions, fixed-reference coverage and retrieval-assisted detection. Its interpretation remains limited to this patient and the complete construction configurations compared.}

From a clinical workflow perspective, the E1--E4 grading scheme is useful because it separates direct contradictions from unsupported claims. \rev{E4 judgments identify explicit conflicts for review, whereas E3 judgments identify absence of retrieved support. Alert prioritization by clinical harm requires independent clinical assessment.} This distinction is important for physician-facing systems because a binary hallucination label may not provide enough information for triage. The structured output also supports auditability \rev{by marking suspicious sentences and indicating whether they lack support or conflict with} explicit patient evidence.

\rev{The Meditron-7B case study illustrates detector predictions on naturally generated summaries; assessing accuracy and generalization requires a pre-specified, independently annotated sample.}

\rev{CuraView combines explicit knowledge representation, graph retrieval, structured evidence grading and model adaptation for patient-specific text verification. Interpretation is limited to the evaluated sample and configuration; generalization to other patients and settings requires validation.}

\section{Limitations and Future Work}

First, the evaluation is based on the Discharge-Me dataset derived from MIMIC-IV records at a single U.S. academic medical center. Although this dataset is widely used in clinical NLP research, documentation practices, coding conventions, clinical workflows, and terminology can vary across institutions and countries. External validation on additional EHR sources, institution-specific discharge datasets, and non-English clinical records is necessary before the system can be considered broadly applicable.

Second, the current verification scope focuses on the brief hospital course portion of discharge summaries. This section contains clinically important information, but discharge quality also depends on medication reconciliation, discharge instructions, follow-up plans, pending tests, and patient education. Extending CuraView to these parts of the discharge document will require broader relation schemas and richer evidence modeling. It may also require integration with medical ontologies and terminology systems such as UMLS, SNOMED-CT, RxNorm, and ICD mappings \cite{wu2025medicalgraphrag}. Such resources can support drug interaction checks, contraindication reasoning, synonym normalization, and guideline-level consistency that may not be explicit in the patient chart.

\rev{Third, labels are derived from controlled hallucination generation. Whether generated perturbations constitute errors, whether unmodified sentences are faithful, and whether E3/E4 assignments are correct have not been verified through independent clinical annotation; errors in the reference labels may affect the reported metrics. Full discharge records in the main experiment may expose the original brief hospital course within the evidence context. The longitudinal-record exploration removes recognized target sections but may retain semantically related content. The main results therefore characterize verification with access to original source content. Verification independent of the original target requires a separate evaluation in which target sections are removed before graph construction and the derived evidence is regenerated.}

\rev{Fourth, error-type annotation and prediction are single-label. Additional errors within a sentence may remain unrepresented by the assigned label. Evaluating the detection of co-occurring error types requires multi-label reference annotations.}

\rev{Fifth, the study selects consecutive local identifiers 1-\/-250 from a single-center benchmark, holding out 201-\/-250 for testing; random splitting of the upstream resources does not ensure that the first 250 local patients are representative.}

\rev{Sixth,} CuraView still produces false negatives and false positives. False negatives can occur when medical synonyms, abbreviations, or paraphrases are not properly aligned with the graph representation. False positives can occur when a clinically valid statement is only implied by the record or appears in free text rather than structured fields. \rev{The supplementary evaluation on unmodified target documents (Section~\ref{sec:unmodified-docs}) showed a high false-positive burden, and 148 of its sentences did not receive a valid decision; these results apply only to the evaluated model configuration and the subset with valid decisions.} Future work should improve synonym normalization, temporal reasoning, uncertainty calibration, and confidence-based triage. \rev{Future model and prompt selection will be performed on validation data, and the selected configuration will be frozen before independent testing.}

\rev{The single-case graph exploration relies on provisional model-assisted reference facts, includes assertion omissions and boundary ambiguity, and compares graph configurations with unequal generation ceilings. The same patient had already been used for exploration, so later results do not constitute independent validation on a new patient. Community reports did not undergo clinical review, and the binary probes do not estimate whole-document false-positive rates. In the exploratory study, the summarization prompt asks the model to resolve conflicting descriptions, which may obscure differences across sources or time points. The community-report templates also retain nonclinical instructions concerning legal compliance and reputational impact. Future work will evaluate clinically focused templates designed to preserve source attribution, timing, uncertainty, and conflicting evidence.}

Finally, deployment requires careful engineering, privacy protection, and clinical governance. Per-patient graph construction introduces computational overhead, and local inference currently assumes GPU resources sufficient for quantized 14B-scale inference. Future work should explore incremental graph updates, model distillation, more compact inference settings, and privacy-preserving deployment. \rev{Clinical harm severity remains unassessed and requires a separate expert-designed evaluation.} Before clinical use, CuraView should be validated under institutional review, integrated with EHR access control policies, and used as a physician-assistive quality-control tool rather than an autonomous decision-making system.

\section{Conclusion}

We introduced CuraView, a knowledge-based multi-agent framework for sentence-level hallucination detection in discharge summaries. CuraView formulates the task as patient-specific claim verification and combines GraphRAG-based EHR knowledge graphs, a seven-type clinical hallucination taxonomy, and an E1--E4 evidence grading scheme. This design allows the system to identify whether a sentence is supported, unsupported, or directly contradicted by patient-specific evidence.

\rev{Against the generation-derived reference labels, with the original discharge record available as a reference, the curated fine-tuned Qwen3-14B verifier has the highest E4 and E3+E4 F1 point estimates among the evaluated CuraView configurations in the fixed patient-split comparison. Comparisons with local RAGTruth-style and QAGS-style methods reflect complete configurations with different retrieval, verification and output procedures and do not isolate GraphRAG's contribution. The separate API exploration uses three admissions from one patient: Domain has higher fixed-reference coverage than Generic and fewer probe-level false positives than Generic and Flat Dense. With unequal generation ceilings and provisional reference facts, these results support only an exploratory comparison of the evaluated configurations in this single case.}

\rev{CuraView integrates} knowledge representation, evidence retrieval, interpretable grading, and multi-agent data curation for high-risk medical text generation. For knowledge-based systems, the study illustrates how patient-level knowledge graphs and LLM-based reasoning can be combined into a transparent verification architecture. Future work will focus on external validation, expert-annotated natural hallucination datasets, richer medical ontology integration, improved calibration, and deployment-oriented compression.

\RevHeading\section{\rev{Acknowledgements}}

\rev{The authors gratefully acknowledge the creators and maintainers of the Discharge-Me dataset and the MIMIC-IV clinical database for making these resources available to the research community. The authors also thank the developers of Qwen3 for open-sourcing the base models used in this work.}

\section{\rev{CRediT authorship contribution statement}}

\rev{Severin Ye: Conceptualization, Methodology, Software, Data curation, Investigation, Validation, Formal analysis, Visualization, Writing -- original draft, Writing -- review \& editing, Project administration. Xiao Kong: Visualization, Writing -- review \& editing. Limei Peng: Writing -- review \& editing. Xiaopeng He: Validation, Formal analysis. Guangsu Yan: Software. Dongsuk Oh: Resources, Supervision, Writing -- review \& editing.}

\section{\rev{Declaration of competing interest}}

\rev{The authors declare that they have no known competing financial interests or personal relationships that could have appeared to influence the work reported in this paper.}

\section{\rev{Funding}}

\rev{This research did not receive any specific grant from funding agencies in the public, commercial, or not-for-profit sectors.}

\section{\rev{Data availability}}

\rev{The study uses publicly available benchmark and dataset resources. Access to and redistribution of original patient-level data are governed by the source datasets' conditions; the authors do not hold or redistribute these data outside those conditions. Supplementary materials distinguish implementation templates, synthetic examples and patient-derived artifacts. Any release of patient-derived training data is also subject to the source data access and redistribution conditions; no public release of patient-level data is claimed for this study.}

\section{\rev{Declaration of generative AI and AI-assisted technologies in the manuscript preparation process}}

\rev{During the preparation of this work, the authors used ChatGPT to assist with manuscript structure refinement, bilingual translation, language polishing, sentence and paragraph rewriting, drafting and revising parts of the abstract and main text, drafting submission-related statements, interpreting revision feedback, and supporting script and formatting adjustments related to manuscript preparation. The authors used ChatGPT to explore initial figure layouts. After using these tools, the authors reviewed and edited all content as needed and take full responsibility for the content of the published article.}

\bibliographystyle{cas-model2-names}
\bibliography{ref}

\input{supplementary_materials}

\end{document}

%% file: supplementary_materials.tex
\clearpage
\begingroup\RevisionInk\RevTable
\appendix
\setcounter{table}{0}
\renewcommand{\thetable}{S\arabic{table}}

\section{Supplementary Materials}
\label{sec:supplement}

\subsection{Entity-type inventories}
\label{sec:S-entity-types}

\begin{center}
\begin{minipage}{\linewidth}
\centering
\captionof{table}{\rev{Entity-type inventories for the main experiment and the exploratory graph comparison.}}
\label{tab:S-entity-types}
\small
\setlength{\tabcolsep}{4pt}
\renewcommand{\arraystretch}{1.15}
\begin{tabular}{>{\raggedright\arraybackslash}p{0.27\textwidth}>{\raggedright\arraybackslash}p{0.10\textwidth}>{\raggedright\arraybackslash}p{0.53\textwidth}}
\toprule
Setting & Number of types & Entity types \\
\midrule
Main experiment, as described in Table~\ref{tab:entity-types} & 9 & PATIENT, DIAGNOSIS, MEDICATION, LAB\_TEST, LAB\_RESULT, VITAL\_SIGN, SYMPTOM, PROCEDURE, DEPARTMENT \\
Exploratory Generic and Domain configurations & 15 & PATIENT, CLINICAL\_EVENT, EXAMINATION, IMAGING\_FINDING, ANATOMICAL\_SITE, DISEASE, SYMPTOM, MEASUREMENT, CLINICAL\_IMPRESSION, TIMEPOINT, DEMOGRAPHIC\_ATTRIBUTE, LAB\_TEST, LAB\_RESULT, TREATMENT, PROCEDURE \\
\bottomrule
\end{tabular}

\vspace{3pt}
\begin{minipage}{0.92\textwidth}
\footnotesize Medications have a dedicated MEDICATION type in Table~\ref{tab:entity-types} and are included under TREATMENT in the exploratory Domain prompt.
\end{minipage}
\end{minipage}
\end{center}

\subsection{Implementation materials and experimental settings}
\label{sec:S-implementation}

The \href{https://github.com/severin-ye/CuraView/tree/d6117ca87f9418860bf4d3e4483aa73eae795bca}{CuraView repository} provides the implementation and prompt templates. The accompanying file \texttt{Table\_Guide.md} lists the materials for each table and the checksums of the Domain extraction prompts.

\let\savedHtable\theHtable
\renewcommand{\theHtable}{supp.\arabic{table}}

\subsubsection{Evaluation and implementation materials}
\label{sec:q12-supp-eval}

\begin{center}
\begin{minipage}{\linewidth}
\centering
\captionof{table}{\rev{Implementation and evaluation materials and their associated analyses.}}
\label{tab:S-materials}
\footnotesize
\setlength{\tabcolsep}{3.5pt}
\renewcommand{\arraystretch}{1.18}
\begin{tabular}{>{\raggedright\arraybackslash}p{0.15\textwidth}>{\raggedright\arraybackslash}p{0.15\textwidth}>{\raggedright\arraybackslash}p{0.38\textwidth}>{\raggedright\arraybackslash}p{0.22\textwidth}}
\toprule
Material & Associated analysis & Contents & Source \\
\midrule
Implementation and dependencies & Section~\ref{sec:implementation} & Separate environments for sentence detection and the exploratory graph comparison. & \href{https://github.com/severin-ye/CuraView/blob/2b9f3fb93dd80462187c46a89613aa8dbf23fe39/graph/README.md}{installation guide}; \href{https://github.com/severin-ye/CuraView/blob/2b9f3fb93dd80462187c46a89613aa8dbf23fe39/detection/README.md}{detector guide} \\
Graph integration test & Section~\ref{sec:longcase} & Synthetic graph construction, retrieval, and comparison across the three conditions. & \href{https://github.com/severin-ye/CuraView/blob/2b9f3fb93dd80462187c46a89613aa8dbf23fe39/graph/test_native.py}{integration test}; \href{https://github.com/severin-ye/CuraView/blob/2b9f3fb93dd80462187c46a89613aa8dbf23fe39/comparison/README.md}{run instructions} \\
Current detection templates & Section~\ref{sec:evidence-grading} & Evidence grades and structured output for sentence verification. & \href{https://github.com/severin-ye/CuraView/tree/2b9f3fb93dd80462187c46a89613aa8dbf23fe39/detection/doc/prompts}{detector templates}; \href{https://github.com/severin-ye/CuraView/blob/2b9f3fb93dd80462187c46a89613aa8dbf23fe39/detection/curaview/_03_detection/parsing/output_contract.py}{output contract} \\
Prompt construction & Section~\ref{sec:implementation} & Detector prompt loading and assembly of comparison requests. & \href{https://github.com/severin-ye/CuraView/blob/2b9f3fb93dd80462187c46a89613aa8dbf23fe39/detection/curaview/_03_detection/prompts/prompt_loader.py}{prompt loader}; \href{https://github.com/severin-ye/CuraView/blob/2b9f3fb93dd80462187c46a89613aa8dbf23fe39/comparison/detection.py}{comparison requests} \\
Generation prompts & Perturbation generation & Recovered prompts for assessing sentence suitability and generating a targeted perturbation. The public copy omits the original examples. & \href{https://github.com/severin-ye/CuraView/tree/d6117ca87f9418860bf4d3e4483aa73eae795bca/generation/prompt}{generation prompts} \\
External baseline input preparation & Table~\ref{tab:new-baselines} & Converts candidate sentences, reference texts and labels into files for baseline evaluation. & \href{https://github.com/severin-ye/CuraView/blob/d6117ca87f9418860bf4d3e4483aa73eae795bca/evaluation/official_baseline_inputs.py}{input conversion script} \\
AlignScore implementation & Table~\ref{tab:new-baselines} & Selects a threshold on the calibration set and applies it to test scores. & \href{https://github.com/severin-ye/CuraView/blob/d6117ca87f9418860bf4d3e4483aa73eae795bca/evaluation/alignscore_calibrated.py}{AlignScore calibration and evaluation} \\
QAGS implementation & Table~\ref{tab:new-baselines} & Runs QAGS with the no-answer option enabled. & \href{https://github.com/severin-ye/CuraView/blob/d6117ca87f9418860bf4d3e4483aa73eae795bca/evaluation/qags_noans_pipeline.py}{QAGS no-answer pipeline} \\
FactCC implementation & Table~\ref{tab:new-baselines} & Scores reference--sentence pairs with the third-party FactCC checkpoint. & \href{https://github.com/severin-ye/CuraView/blob/d6117ca87f9418860bf4d3e4483aa73eae795bca/evaluation/run_factcc_hf_candidate.py}{FactCC evaluation script} \\
External baseline evaluation &
Table~\ref{tab:new-baselines} &
We calculated F1 as the harmonic mean of patient-averaged precision and recall. We recalculated F1 in 2,000 patient-level bootstrap samples to obtain 95\% percentile confidence intervals. &
Section~\ref{sec:evaluation-metrics} and Table~\ref{tab:new-baselines}. \\
\bottomrule
\end{tabular}
\end{minipage}
\end{center}

The current detection templates describe the revised method specification.

\paragraph{Offline software checks.}
The sentence detector uses GraphRAG 2.7, and the exploratory graph comparison uses GraphRAG 3.3. The \href{https://github.com/severin-ye/CuraView/blob/2b9f3fb93dd80462187c46a89613aa8dbf23fe39/graph/README.md}{installation guide} gives the dependencies and commands for each environment. From the repository root, the following command scores four synthetic sentences:
\begin{verbatim}
python evaluation/synthetic_check.py
\end{verbatim}
The expected precision, recall, and F1 are 1.0, 0.5, and 0.667, respectively. The graph integration test in Table~\ref{tab:S-materials} checks graph construction and retrieval on synthetic documents.

\paragraph{External baseline evaluation.}
The external baselines use 1,103 test sentences from 50 patients and a separate calibration set of 1,149 sentences from 50 patients. Thresholds are selected on the calibration set. Confidence intervals are calculated using 2,000 patient-cluster bootstrap replicates, sampling patients with replacement and retaining multiplicity.

AlignScore uses the official AlignScore-base checkpoint in \texttt{nli\_sp} mode. QAGS uses the no-answer variant with \texttt{use\_no\_ans=true} and \texttt{use\_only\_no\_ans=false}. FactCC uses the \texttt{manueldeprada/FactCC} model with a maximum input length of 512 tokens and \texttt{only\_first} truncation.

\paragraph{Paired comparison.}
The archived predictions covered 709 of 710 reference sentences for Curated FT (32 patients), 1,132 of 1,138 for Base (50 patients), and all 1,103 for AlignScore (50 patients). After matching sentence hashes and excluding missing predictions and known text mismatches, we retained 616 sentences, of which 226 had positive reference labels. For Curated FT and Base, we classified E3 and E4 predictions as positive. For 32 matched sentences, the binary decision recorded by Base differed from the classification based on its evidence grade. Using the recorded binary decisions gave a Base F1 of 0.812 and a Curated FT--Base F1 difference of 0.049 (95\% CI \mbox{0.015--0.085}).

\subsubsection{Recovered training materials}

Table~\ref{tab:S-training} reports the settings specified in the recovered Qwen3-14B-Base training script, which uses MS-SWIFT and LoRA. These settings describe the recovered script; its correspondence to the checkpoints used for the main results is not established by the available records.

\begin{center}
\begin{minipage}{\linewidth}
\centering
\captionof{table}{\rev{Settings specified in the recovered training script.}}
\label{tab:S-training}
\small
\setlength{\tabcolsep}{6pt}
\renewcommand{\arraystretch}{1.12}
\begin{tabular}{ll}
\toprule
Parameter & Value \\
\midrule
Learning rate & $5\times10^{-5}$ \\
Configured epochs & 2 \\
LoRA rank & 16 \\
LoRA alpha & 32 \\
Per-device batch size & 1 \\
Gradient accumulation & 8 steps \\
Maximum sequence length & 12,288 tokens \\
Learning-rate schedule & Cosine \\
Warmup ratio & 0.03 \\
Weight decay & 0.01 \\
Validation split ratio & 0.05 \\
\bottomrule
\end{tabular}
\end{minipage}
\end{center}

LoRA target modules are selected with \texttt{find\_all\_linears(model)}. The \href{https://github.com/severin-ye/CuraView/blob/d6117ca87f9418860bf4d3e4483aa73eae795bca/training/train_qwen3_14b_lora.py}{training script} specifies the settings reported in Table~\ref{tab:S-training}.

\subsubsection{Training--validation split documentation}
\label{sec:S-split}

The recovered training script specifies a validation fraction of 0.05 and a data seed of 42. Splitting is delegated to MS-SWIFT without an explicit patient or source--variant grouping rule. Training and validation membership for the reported model is not available in the retained materials; overlap by patient or source--variant pair therefore cannot be quantified.

\subsubsection{Exploratory graph and detection materials}

The Generic and Domain configurations use the same patient records, chunking, and 15 entity types. Their output-token ceilings are 8,192 and 32,768, respectively. Domain also includes one completed extraction from the earlier 8,192-token configuration. The comparison therefore evaluates the two graph-construction configurations.

\begin{center}
\begin{minipage}{\linewidth}
\centering
\captionof{table}{\rev{Materials for the exploratory graph and detection comparison.}}
\label{tab:S-exploratory}
\footnotesize
\setlength{\tabcolsep}{3.5pt}
\renewcommand{\arraystretch}{1.18}
\begin{tabular}{>{\raggedright\arraybackslash}p{0.15\textwidth}>{\raggedright\arraybackslash}p{0.16\textwidth}>{\raggedright\arraybackslash}p{0.37\textwidth}>{\raggedright\arraybackslash}p{0.22\textwidth}}
\toprule
Material & Associated analysis & Contents & Source \\
\midrule
Generic templates & Tables~\ref{tab:prompt-optimization} and~\ref{tab:longcase-quality} & Resolved graph-construction templates. & \href{https://github.com/severin-ye/CuraView/tree/2b9f3fb93dd80462187c46a89613aa8dbf23fe39/graph/v33/prompts}{graph templates} \\
Domain templates & Tables~\ref{tab:prompt-optimization} and~\ref{tab:longcase-quality} & Chinese extraction prompt used for the reported Domain graph and English translation used by the public implementation. & \href{https://github.com/severin-ye/CuraView/blob/d6117ca87f9418860bf4d3e4483aa73eae795bca/graph/v33/prompts/domain.extract.txt}{Chinese prompt}; \href{https://github.com/severin-ye/CuraView/blob/2b9f3fb93dd80462187c46a89613aa8dbf23fe39/graph/v33/prompts/domain.extract.txt}{English translation} \\
Comparison interface & Table~\ref{tab:longcase-detection} & Flat Dense, Generic, and Domain retrieval conditions and the detector interface. & \href{https://github.com/severin-ye/CuraView/blob/2b9f3fb93dd80462187c46a89613aa8dbf23fe39/comparison/paired.py}{comparison code}; \href{https://github.com/severin-ye/CuraView/blob/2b9f3fb93dd80462187c46a89613aa8dbf23fe39/comparison/README.md}{run instructions} \\
Exploratory results & Tables~\ref{tab:longcase-quality} and~\ref{tab:longcase-detection} & Graph agreement with source records and detection outcomes across three retrieval conditions. & Tables~\ref{tab:longcase-quality} and~\ref{tab:longcase-detection}. Supplementary Data S1--S3 provide the graph review records. \\
\bottomrule
\end{tabular}
\end{minipage}
\end{center}

The detection comparison uses \texttt{qwen-plus} with a 6,000-character evidence limit and 12 probes from one patient: six source-supported statements and six direct contradictions. We used the Chinese extraction prompt to build the Domain graph evaluated in Tables~\ref{tab:prompt-optimization} and~\ref{tab:longcase-quality}. The \href{https://github.com/severin-ye/CuraView/blob/2b9f3fb93dd80462187c46a89613aa8dbf23fe39/graph/README.md}{public comparison release} uses the English translation. All three retrieval conditions in Table~\ref{tab:longcase-detection} used the same \href{https://github.com/severin-ye/CuraView/blob/2b9f3fb93dd80462187c46a89613aa8dbf23fe39/comparison/detection_prompt.md}{detector prompt} and \href{https://github.com/severin-ye/CuraView/blob/2b9f3fb93dd80462187c46a89613aa8dbf23fe39/comparison/strategy.txt}{verification strategy}.

\subsection{Detection implementation materials}
\label{sec:S-detection}

\begin{center}
\begin{minipage}{\linewidth}
\centering
\captionof{table}{\rev{Detection implementation materials.}}
\label{tab:S-detection-materials}
\footnotesize
\setlength{\tabcolsep}{4pt}
\renewcommand{\arraystretch}{1.18}
\begin{tabular}{>{\raggedright\arraybackslash}p{0.17\textwidth}>{\raggedright\arraybackslash}p{0.39\textwidth}>{\raggedright\arraybackslash}p{0.34\textwidth}}
\toprule
Material & Files & Purpose \\
\midrule
Detection prompts & \href{https://github.com/severin-ye/CuraView/tree/2b9f3fb93dd80462187c46a89613aa8dbf23fe39/detection/doc/prompts}{detector templates} & Current sentence verification templates and output requirements for API and local inference \\
Output schema & \href{https://github.com/severin-ye/CuraView/blob/2b9f3fb93dd80462187c46a89613aa8dbf23fe39/detection/curaview/_03_detection/validation/schemas.py}{\texttt{schemas.py}} & Define output fields, data types, and permitted values \\
Local inference implementation & \href{https://github.com/severin-ye/CuraView/blob/2b9f3fb93dd80462187c46a89613aa8dbf23fe39/detection/curaview/backends/local/local_llm.py}{\texttt{local\_llm.py}} & Load the local prompt and assemble the structured prediction \\
Scoring implementation & \href{https://github.com/severin-ye/CuraView/blob/d6117ca87f9418860bf4d3e4483aa73eae795bca/evaluation/reports/compare_systems.py}{\texttt{compare\_systems.py}}; \href{https://github.com/severin-ye/CuraView/blob/d6117ca87f9418860bf4d3e4483aa73eae795bca/evaluation/reports/tools/metrics_calculator.py}{\texttt{metrics\_calculator.py}} & Calculate detection metrics from reference and prediction records \\
Supplementary table guide & \texttt{Table\_Guide.md} & Identify the materials for each analysis and the checksums of the Domain extraction prompts \\
\bottomrule
\end{tabular}

\vspace{3pt}
\begin{minipage}{0.92\textwidth}
\footnotesize The updated templates specify the current grading rules. The reported evaluations were not rerun with these templates. The available records do not identify the exact prompt version used for Table~\ref{tab:q12-e34}.
\end{minipage}
\end{minipage}
\end{center}
\par
The \href{https://github.com/severin-ye/CuraView/blob/2b9f3fb93dd80462187c46a89613aa8dbf23fe39/supplement/Executed_Example.md}{worked example} pairs an excerpt from the retrieved evidence with the complete JSON response from the exploratory \texttt{qwen-plus} detector. Figure~\ref{fig:structured-output} uses \texttt{hallucination\_status} for the API field \texttt{is\_hallucination}, preserving its Boolean value. A separate synthetic example provides the complete request, call parameters, and expected output.
\let\theHtable\savedHtable

\subsection{Baseline implementations and calibration}
\label{sec:S-baselines}

The comparison includes two local clinical verification workflows and three external factual-consistency methods. Table~\ref{tab:S-baselines} summarizes their roles, implementations, and decision procedures. The local QAGS-style workflow and the external QAGS no-answer variant are distinct implementations.

\begin{center}
\begin{minipage}{\linewidth}
\centering
\captionof{table}{\rev{Baseline implementations and calibration procedures.}}
\label{tab:S-baselines}
\scriptsize
\setlength{\tabcolsep}{3pt}
\renewcommand{\arraystretch}{1.2}
\begin{tabular}{>{\raggedright\arraybackslash}p{0.12\textwidth}>{\raggedright\arraybackslash}p{0.14\textwidth}>{\raggedright\arraybackslash}p{0.31\textwidth}>{\raggedright\arraybackslash}p{0.19\textwidth}>{\raggedright\arraybackslash}p{0.14\textwidth}}
\toprule
Baseline & Comparison role & Implementation and evidence processing & Output and decision rule & Adaptation for this study \\
\midrule
RAGTruth-style EHR & Flat retrieval with LLM-based verification & Local literature-inspired workflow that retrieves flat EHR passages and uses Qwen3-Plus to assess consistency & Sentence-level binary consistency judgment & Local EHR verification workflow \\
QAGS-style EHR & Question-based verification & Local workflow that generates questions from candidate sentences, answers them using EHR source text, and compares the answers & Sentence-level binary decision based on answer consistency & Local EHR question-generation and answering workflow \\
AlignScore & Alignment-based factual-consistency scoring & Official implementation and AlignScore-base checkpoint in \texttt{nli\_sp} mode; context is grouped at sentence boundaries, with maximum support across chunks for each claim sentence and subsequent averaging across claim sentences & Alignment support score; hallucination if $s < 0.7167296707630157$ & Pretrained checkpoint with threshold calibration \\
QAGS (no-answer variant) & Question-answer consistency scoring & Authors' question-generation and QA model resources; reference text is processed by the authors' QA pipeline through the local no-answer-enabled UTF-8 implementation & Answer-overlap F1 score; hallucination if $s < 0.8366666666666667$ & Pretrained model resources with threshold calibration \\
FactCC (third-party checkpoint) & Classification-based factual-consistency scoring & \texttt{manueldeprada/FactCC}; reference--sentence input pairs, a maximum length of 512 tokens, and \texttt{truncation=only\_first} for the reference text & Probability of the CORRECT class; hallucination if $s < 0.9992513358592987$ & Pretrained checkpoint with threshold calibration \\
\bottomrule
\end{tabular}

\vspace{3pt}
\begin{minipage}{0.95\textwidth}
\footnotesize Notes: For the three external baselines, higher scores indicate greater factual consistency; scores at or above the threshold produce a non-hallucination prediction. Model weights remain fixed during calibration. The local workflow descriptions follow the manuscript. Complete run configurations for the two local workflows and run-specific QA-window settings for the external QAGS variant are absent from the retained materials.
\end{minipage}
\end{minipage}
\end{center}

\paragraph{Calibration protocol.}
The calibration set contained 1,149 sentences from 50 patients, comprising 739 unmodified sentences and 410 generated perturbations. These patients belonged to the study's non-test patient pool and had no patient-identifier overlap with the test set. Threshold candidates consisted of 0, 1, and the midpoints between distinct calibration scores. The selected threshold maximized the unweighted mean of per-patient hallucination F1. Patient-level F1 was set to zero when \(2TP+FP+FN=0\), and ties were resolved by selecting the smallest threshold. Test labels were not used for threshold selection.

\paragraph{Comparison scope.}
Med-HALT \cite{pal2023medhalt} evaluates reasoning and knowledge recall in medical question answering, whereas RadFlag \cite{zhang2024radflag} assesses consistency across repeated radiology-report generations. Their original protocols differ from verification of fixed candidate sentences against patient records. The reported comparisons cover the general-purpose factual-consistency methods and local clinical verification workflows listed in Table~\ref{tab:S-baselines}.

\subsection{Sampling and denominators for the graph assessment}
\label{sec:S-graph-sampling}

GPT-6-Astra extracted assertions from anonymized graph records. Initial source-support assessment was conducted in five batches, followed by adjudication in a separate GPT-6.1-Sol context. The adjudicator received the initial judgments and supporting evidence, but not the graph identity key. This review did not include independent clinician adjudication.

Each graph contributed 20 entity-description records and 40 relationship records. Atomization separated statements that could be assessed individually against the source, and repeated assertions within each graph were counted once. Two Generic and nine Domain records yielded no eligible clinical assertions and were excluded from record-level fidelity. The remaining 58 Generic records yielded 138 record--assertion associations and 137 unique assertions after deduplication. The remaining 51 Domain records yielded 64 associations and 63 unique assertions. Table~\ref{tab:S-graph-sampling} summarizes these counting units.

For example, a Generic record containing three laboratory results and one interpretation yielded four assertions. A Domain record containing only a test name, without a result or event, yielded no eligible assertion. The supplementary data provide the record-to-assertion mappings, judgments before and after adjudication, and findings from the prespecified review subset.

\begin{center}
\begin{minipage}{\linewidth}
\centering
\captionof{table}{\rev{Sampling and counting units for the graph assessment.}}
\label{tab:S-graph-sampling}
\small
\setlength{\tabcolsep}{5pt}
\renewcommand{\arraystretch}{1.12}
\begin{tabular}{>{\raggedright\arraybackslash}p{0.58\textwidth}rr}
\toprule
Measure & Generic & Domain \\
\midrule
Sampled entity-description records & 20 & 20 \\
Sampled relationship records & 40 & 40 \\
Metadata-only records excluded from record-level fidelity & 2 & 9 \\
Records retained for record-level fidelity & 58 & 51 \\
Unique assertions used for source-support assessment & 137 & 63 \\
Supported assertions & 70 & 57 \\
Records in the prespecified review subset & 12 & 12 \\
Records with assertion-boundary or omission issues in that subset & 2 & 3 \\
\bottomrule
\end{tabular}

\vspace{3pt}
\begin{minipage}{0.80\textwidth}
\footnotesize Metadata-only exclusions comprised 0 entity-description records and 2 relationship records for Generic, and 4 entity-description records and 5 relationship records for Domain. Contradicted, unsupported, and unresolved assertions remained in the source-support denominator. The final row counts affected records, not individual assertions.
\end{minipage}
\end{minipage}
\end{center}

\begin{center}
\begin{minipage}{\linewidth}
\centering
\captionof{table}{\rev{Findings from the prespecified record review and their treatment in the reported assertion inventories.}}
\label{tab:S-graph-review}
\small
\setlength{\tabcolsep}{4pt}
\renewcommand{\arraystretch}{1.18}
\begin{tabular}{>{\raggedright\arraybackslash}p{0.17\textwidth}>{\raggedright\arraybackslash}p{0.36\textwidth}>{\raggedright\arraybackslash}p{0.37\textwidth}}
\toprule
Graph and record & Review finding & Treatment in the reported inventory \\
\midrule
Generic, e499 & Location and examination purpose were counted as assertions & Support retained for these statements \\
Generic, e550 & Assessment targets were counted without examination findings & Support retained for the stated assessment targets \\
Domain, e1065 & A correct value and an incorrect unit qualifier were combined & The combined assertion remained contradicted \\
Domain, r084 & Findings from a linked node were omitted & Support retained for the documented examination; omitted findings were not added \\
Domain, r194 & A truncated expression was interpreted as a threshold & Support retained only for the literal statement \\
\bottomrule
\end{tabular}
\end{minipage}
\end{center}

The five flagged records remained in the reported inventories. Separately, adjudication changed assertion A-0059 from supported to unresolved because the source did not establish the stated temporal association. Generic support therefore decreased from 71/137 (51.8\%) to 70/137 (51.1\%), while Domain support remained at 57/63 (90.5\%). These rates use the retained assertion inventories.

\endgroup